\documentclass[10pt,twocolumn,letterpaper,table]{article}

\usepackage[pagenumbers]{cvpr} 

\usepackage{amsmath,amsfonts,bm}

\def\eqref#1{equation~\ref{#1}}
\def\Eqref#1{Equation~\ref{#1}}

\def\1{\bm{1}}

\def\rh{{\textnormal{h}}}

\def\rvm{{\mathbf{m}}}

\def\rvp{{\mathbf{p}}}

\def\rvx{{\mathbf{x}}}

\def\ervb{{\textnormal{b}}}

\def\ervl{{\textnormal{l}}}

\def\ervo{{\textnormal{o}}}
\def\ervp{{\textnormal{p}}}

\def\ervs{{\textnormal{s}}}

\def\vmu{{\bm{\mu}}}

\def\vm{{\bm{m}}}

\def\vp{{\bm{p}}}

\def\vr{{\bm{r}}}
\def\vs{{\bm{s}}}

\def\vv{{\bm{v}}}

\def\vx{{\bm{x}}}

\def\evmu{{\mu}}

\def\evsigma{{\sigma}}

\def\evl{{l}}

\def\evp{{p}}

\def\evs{{s}}

\def\evv{{v}}
\def\evw{{w}}

\def\mH{{\bm{H}}}
\def\mI{{\bm{I}}}

\DeclareMathAlphabet{\mathsfit}{\encodingdefault}{\sfdefault}{m}{sl}
\SetMathAlphabet{\mathsfit}{bold}{\encodingdefault}{\sfdefault}{bx}{n}

\newcommand{\E}{\mathbb{E}}

\DeclareMathOperator*{\argmax}{arg\,max}
\DeclareMathOperator*{\argmin}{arg\,min}

\usepackage{soul}
\usepackage{comment}
\usepackage{xfrac}
\usepackage{multirow}
\usepackage{overpic}
\usepackage{amssymb,amsthm}

\newcommand{\tr}[1]{{\color{red} #1}}
\newcommand{\tb}[1]{{\color{blue} #1}}

\usepackage[dvipsnames]{xcolor}

\definecolor{cvprblue}{rgb}{0.21,0.49,0.74}
\usepackage[pagebackref,breaklinks,colorlinks,citecolor=cvprblue]{hyperref}

\usepackage{algorithm,listings}
\newtheorem{proposition}{Proposition}
\newtheorem{lemma}{Lemma}

\def\confName{CVPR}
\def\confYear{2024}

\title{On the Calibration of Human Pose Estimation}

\author{Kerui Gu\textsuperscript{*}\qquad Rongyu Chen\thanks{Equal contribution} \qquad Angela Yao\\
National University of Singapore\\
{\tt \small \{keruigu, rchen, ayao\}@comp.nus.edu.sg
}
}

\begin{document}
\maketitle
\begin{abstract}
Most 2D human pose estimation frameworks estimate keypoint confidence in an ad-hoc manner, using heuristics such as the maximum value of heatmaps. The confidence is part of the evaluation scheme, \eg, AP for the MSCOCO dataset, yet has been largely overlooked in the development of state-of-the-art methods.  
This paper takes the first steps in addressing miscalibration in pose estimation.
From a calibration point of view, the confidence should be aligned with the pose accuracy. In practice, existing methods are poorly calibrated. We show, through theoretical analysis, why a miscalibration gap exists and how to narrow the gap. Simply predicting the instance size and adjusting the confidence function gives considerable AP improvements.  
Given the black-box nature of deep neural networks, however, it is not possible to fully close this gap with only closed-form adjustments. As such, we go one step further and 
learn network-specific adjustments by enforcing consistency between confidence and pose accuracy.  
Our proposed Calibrated ConfidenceNet (CCNet) is a light-weight post-hoc addition that improves AP by up to 1.4\% on off-the-shelf pose estimation frameworks. Applied to the downstream task of mesh recovery, CCNet facilitates an additional 1.0mm decrease in 3D keypoint error. Project page: \\ \href{https://www.comp.nus.edu.sg/~keruigu/calibrate_pose/project.html}{comp.nus.edu.sg/keruigu/calibrate\_pose/project.html}.
\end{abstract}    
\section{Introduction}
\label{sec:intro}

\begin{figure}[t]
  \centering
   \includegraphics[width=0.95\linewidth]{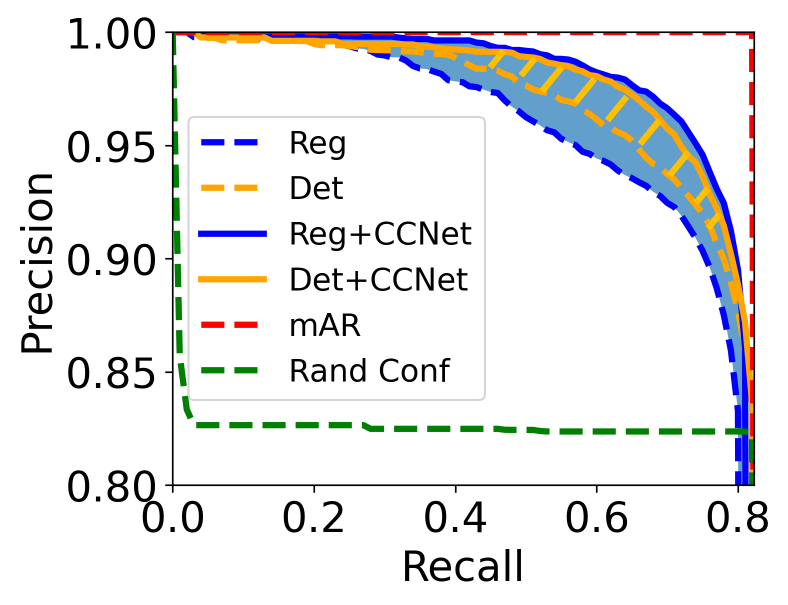}
   \caption{Confidence estimation plays a significant role in the Precision-Recall curve. The area under each curve represents the primary metric AP. As shown, adding CCNet on top of existing detection- and regression-based methods improves the AP. The orange dashed line area and blue area denote the improvement of detection- and regression-based methods, respectively.}
   \label{fig:teaser}
\end{figure}

Two common frameworks for human pose estimation are either based on heatmaps or direct regression.  Heatmap methods~\cite{xiao2018simple, sun2019deep, xu2022vitpose} estimate a non-parametric spatial likelihood on the keypoints in the form of a dense heatmap. Direct regression methods regress the keypoint location either as a deterministic output~\cite{toshev2014deeppose, sun2018integral, gu2021removing} or as a distribution~\cite{li2021human, mao2022poseur}.

The accuracy of the predicted pose is typically based on some distance measure with respect to the ground truth pose. Several metrics for accuracy exist, including end-point-error (EPE) and percentage of correct keypoints (PCK), but these have drawbacks due to their inability to account for the person size and annotation error differences among keypoints, leading to the dominance of Object Keypoint Similarity (OKS). By considering these factors, OKS is perceptually meaningful and a good interpretation of similarity~\cite{ruggero2017benchmarking}.

A more comprehensive and widely accepted metric for evaluating pose is the mean Average Precision (mAP) based on the object keypoint similarity (OKS)~\cite{lin2014microsoft}. This is fashioned directly after object detection evaluation schemes by viewing the pose estimation as a keypoint ``detection''. The mAP, as it is defined, goes beyond evaluating accuracy and also incorporates the corresponding confidence of the estimated pose. A simple experiment shows for SBL~\cite{xiao2018simple} that if we randomize the confidence estimation, the area under curve significantly drops (green dashed line in~\cref{fig:teaser}) and mAP drops from 72.4 to 67.5; instead, if we assign the confidence with the OKS, the area increases (red dashed line in~\cref{fig:teaser}) and mAP increases to 75.6 (see~\cref{sec:motivation} for details). This clearly demonstrates that confidence has a significant impact on evaluation, but is currently overlooked.

First and foremost, the natural question that arises is: how calibrated are current pose estimation frameworks since most approaches use some heuristics such as taking the maximum value of the predicted heamaps~\cite{xiao2018simple, sun2019deep, xu2022vitpose} or uncertainty~\cite{li2021human, mao2022poseur}?  To answer this question, we first analyze, from a statistical point of view, the expected confidence of pose estimation methods versus the ideal confidence given by the OKS under a common assumption that the ground truth annotations follow a Gaussian distribution.
Our analysis reveals systematic miscalibrations based on the way mAP is an exponential envelope functioned by end-point error, instance size, and keypoint annotation falloff. This manifests as a scaling gap for heatmap methods and a form gap for RLE-based methods. Empirically, we verify our analysis by provide a closed-form solution to align the expected OKS, which improves AP with only additionally predicting instance size and changing the confidence form.

However, only correcting the confidence form is not enough since the above analysis is conditioned on a perfect network, which assumes the predicted keypoint can always be located at the center of ground truth distributions. In practice, the network will vary depending on different backbones and datasets, so this assumption is hard to achieve and we aim to learn a network-specific adjustment to better calibrate the confidence.

In this way, we propose a simple yet effective calibration branch, Calibrated ConfidenceNet (CCNet), which is applicable to any pose estimation methods. Specifically, fixing the trained pose models, we base our calibration branch on the penultimate features of the original pose models, which contain rich pose-related information of the input image, and explicitly output score and visibility. We simply supervise the outputs of the calibration branch with the computed OKS and ground truth visibility. By doing so, the predicted confidence directly links to the practical OKS which addresses the issues of different confidence forms and different network characteristics simultaneously. With only a few epochs and negligible additional parameters, the network is much better calibrated and achieves better mAP on various pose models and various datasets. 

Since the predictions from the 2D pose models serve as prior knowledge for downstream tasks, we are among one of the first methods that test the influence of confidence on downstream tasks. We take the example of 3D mesh recovery which fits a SMPL~\cite{loper2023smpl} model to human mesh according to 2D predictions. Experimentally, we show that with a better calibrated pose model, the 2D predictions can produce better 3D results.

Our contribution can be summarized as,
\begin{itemize}
    \item We are the first to study the calibration of 2D pose estimation, which is overlooked in the literature but significantly counts in the evaluation of pose estimation models and the challenging downstream tasks.
    \item We mathematically formulate the ideal form of pose confidence and reveal a mismatch between this and the practical confidence form from current pose estimation methods. A simple solution is provided to verify and refine the misalignment.
    \item We propose a simple but effective method to explicitly model the calibration with minor addition of parameters and training time. Experiments show that adding the calibration branch gives significant improvement on the primary metric mAP and also benefits the downstream tasks.
\end{itemize}

\section{Related Work}\label{sec:related_work}

\paragraph{Pose Estimation.} The past literature in 2D top-down based pose estimation mainly focused on how to improve the accuracy. However, only few works give some heuristic or empirical understandings regarding the pose confidence. Papandreou~\etal~\cite{papandreou2017towards} proposed a re-scoring strategy based on the detected bounding box, which is shown to be effective and applied in most of the following top-down based works~\cite{xiao2018simple, li2021human}. PETR~\cite{shi2022end} empirically found that changing the matching objective to OKS-based improves the AP under the same AR, indicating a better ranking over the samples. Poseur~\cite{mao2022poseur} following regression paradigm noticed previous regression scoring is heuristic and they rescore into likelihood presented by detection maxvals. Although there exist several works that try to change the form of confidence, they remain at the empirical level and lack actual understanding of the confidence. In our paper, we give theoretical understanding of the confidence for both heatmap- and RLE-based methods and analyze their calibration to OKS accordingly. As a result, we propose the Calibrated ConfidenceNet as the solution to achieve better calibration for pose models.

\paragraph{Confidence Estimation} 
is essential in real-world applications and has pushed a lot of work in that direction~\citep{kendall2017uncertainties,lakshminarayanan2017simple,amini2020deep}. Conventional calibration is broadly discussed in the classification~\citep{guo2017calibration} and regression~\citep{song2019distribution}. It reflects the reliability of the confidence to indicate accuracy. The confidence in the context is interpreted as the probability of the prediction being correct. For instance, in \text{classification}, the softmax confidence is viewed as the probability of the image belonging to the class.
Well-calibrated classifier is expected to have confidence approximate accuracy~\cite{guo2017calibration}.
Similarly, recent work~\citep{pathiraja2023multiclass} introduced calibration to boundary box object classification in object detection. When it comes to \text{regression} the definition is vague, among which a quantile-based definition is common~\citep{song2019distribution}.
Evaluation becomes nontrivial in high dimensions. 
However, there is few work studying calibration in pose estimation~\citep{bramlage2023plausible,pierzchlewicz2022multi}. We argue that pose confidence is useful and informative only when it first aligns well with true OKS, otherwise, it cannot be a helpful forecast~\citep{kuleshov2022calibrated}. mAP exactly evaluates this.
\begin{figure*}[t]
    \centering
    \includegraphics[width=0.95\linewidth]{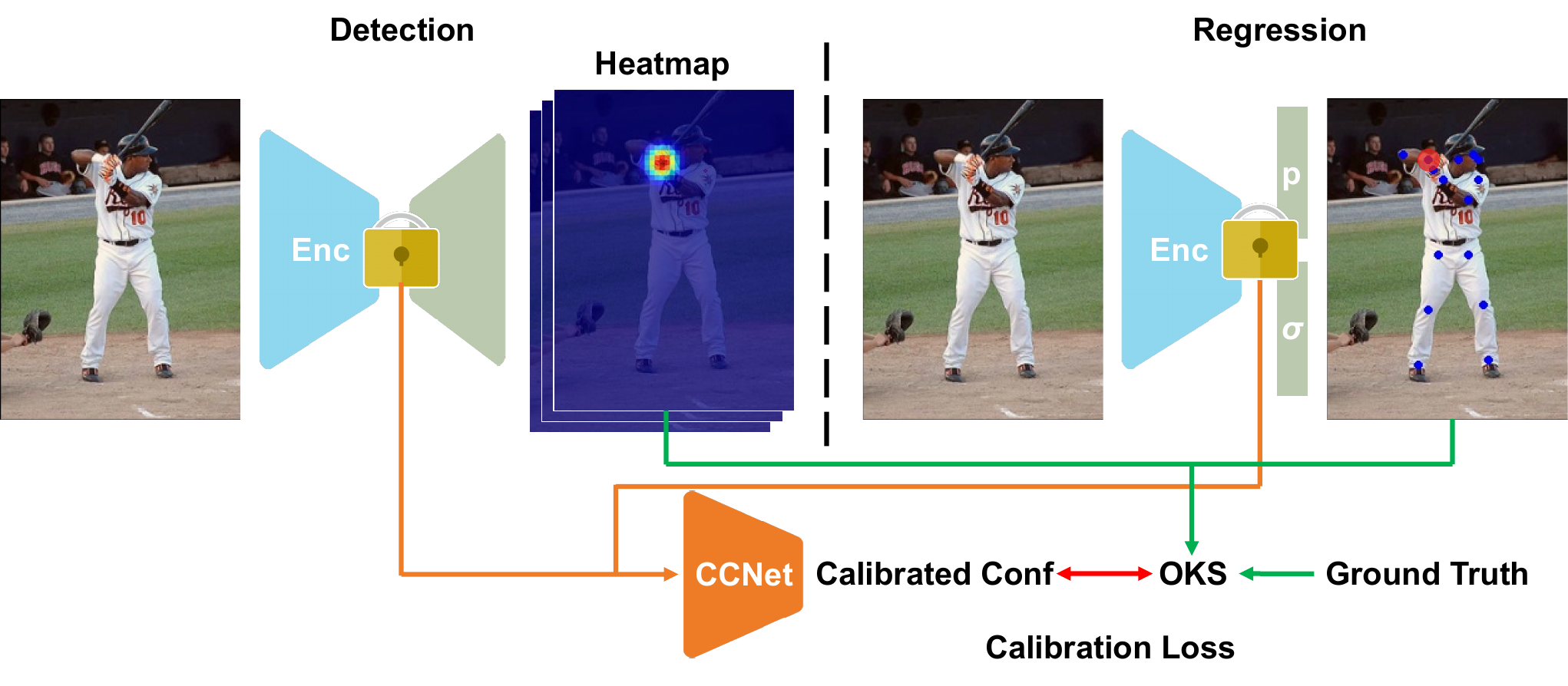}
    \caption{Our calibration method is a post-hoc addition to off-the-shelf pose estimation methods.  CCNet estimates more calibrated confidences directly from latent pose representations and leads to improvement in the mAP.}
    \label{fig:ov}
\end{figure*}

\section{Preliminaries}

\subsection{Human Pose Estimation } 
We consider top-down 2D human pose estimation, where people are already localized and cropped from the scene. Given a single-person image $\mathbf{x}$, pose estimation methods estimate $K$ keypoint coordinates $\hat{\vp}\in\mathbb{R}^{K\times 2}$ and confidence score $\hat{\vs}\in[0,1]^K$.  The keypoint scores are aggregated into a person- or instance--wise confidence $\hat{c} \in [0,1]$ where higher values indicate higher confidence.

\paragraph{Heatmap} methods~\cite{xiao2018simple, sun2019deep, xu2022vitpose} estimate $K$ heatmaps $\hat{\mH}\in\mathbb{R}^{K\times H\times W}$ to represent pseudo-likelihoods, \ie unnormalized probabilities of each pixel being the $k$-th keypoint (Fig.~\ref{fig:ov} {(a)}). The heatmap $\hat{\mH}_k$ can be decoded into the joint coordinate $\hat{\vp}_k$ and confidence $\hat{\evs}_k$ with a simple $\argmax$:
\begin{equation}\label{eq:pre_heatmap}
    \hat{\vp}_k = \mathrm{argmax}(\hat{\mH}_k), 
    \quad \hat{\evs}_k=\max(\hat{\mH}_k),
\end{equation}
\noindent
although more complex forms of decoding have been proposed~\cite{zhang2020distribution} in place of the $\argmax$.

Methods which estimate heatmaps are learned with an MSE loss with respect to a ground truth heatmap $\mH_k$
\begin{equation}\label{eq:detection_loss}
\small
    \mathcal{L}_{\text{det}} = \sum_{k=1}^K\mathrm{MSE}(\hat{\mH}_k, \mH_k)=\sum_{k=1}^K\sum_{i=1}^H\sum_{j=1}^W(\hat{h}_{kij}-h_{kij})^2,
\end{equation}
\noindent
where $(i,j)$ are spatial indices, and $\hat{h}_{kij},h_{kij}$ denotes the $(i,j)^{\text{th}}$ entry of $\hat{\mH}_k$ and $\mH_k$,
Typically, the ground truth heatmap $\mH_k$ is constructed as 2D Gaussian centered at the ground truth keypoint location and a fixed standard deviation $\tilde{l}$, \eg, $\tilde{l}\!=\!2$ for input size $256\times 192$.

\noindent\textbf{Regression} methods directly regress deterministic coordinates of the keypoints or likelihood distributions of the coordinates.
We focus on the state-of-the-art RLE regression~\cite{li2021human, mao2022poseur}, which 
{generally model the likelihood as a distribution (\eg, Normal, Laplace and Normalizing Flows) parameterized by mean and variance parameters $\hat{\vmu}$ and $\hat{\sigma}$.}
Similarly, we can achieve the keypoint prediction and its confidence as
\begin{equation}\label{eq:pre_rle}
    \hat{\rvp}=\hat{\vmu}, \quad \hat{\evs}_k=1-\hat{\sigma}.
\end{equation}

The loss for 
formulated as a Negative Log-Likelihood:
\begin{align}
    \mathcal{L}_{\text{reg}}=-\sum_{k=1}^K\log\hat{p}(\vp_k|\rvx;\hat{\vp}_k,\hat{\sigma}_k),\label{eq:nll}
\end{align}
which can be further expanded as an adaptive weighted error loss between $\hat{\rvp}_k$ and $\rvp_k$, plus a regularization such as $\log\hat{\sigma}_k^2$ in practice.

\noindent \textbf{Other regression-based} methods use heatmap maximum~\citep{wei2020point} or keypoint classification confidence from another head~\citep{li2021pose} or simply fill the confidence as 1~\cite{sun2018integral}. 

\noindent \textbf{Instance-wise confidence scores} are derived by aggregating the keypoint confidences with a weighted summation:
\begin{equation}
    \hat{c}=\mathrm{agg}(\hat{\vs})=\sum_{k=1}^K\hat{\evw}_k\hat{\evs}_k,\text{ where }\hat{\evw}_k=\frac{\mathcal{I}(\hat{\evs}_k>\tau_{\hat{\evs}})}{\sum_{k=1}^K\mathcal{I}(\hat{\evs}_k>\tau_{\hat{\evs}})},\label{eq:thres_vis_weight}
\end{equation}
\noindent
where $\mathcal{I}$ is an indicator function. The convention of the
keypoint-to-instance aggregation $\mathrm{agg}(\cdot)$ is an average function of only selected keypoint predictions with $\hat{\evs}_k>\tau_{\hat{\evs}}$.

\subsection{Pose Estimation Evaluation}
Several metrics have been proposed for evaluating keypoint accuracy, including end-point error (EPE), Percentage of Correct Keypoint (PCK) and Object Keypoint Similarity (OKS)~\cite{lin2014microsoft}.  Of these three error measures, OKS is the most comprehensive.  EPE does not account for the scale of the person; PCK, while normalized with respect to the head size, does not account for keypoint difference. 
OKS factors in both instance size and keypoint variation as an instance measure.  It is defined as
a weighted sum of the exponential envelope of a scaled end-point error:
\begin{align}
&c  =\sum_{k=1}^K\evw_k\exp\left(-\frac{\|\hat{\vp}_k-\vp_k\|^2}{2\evl_k^2}\right),\label{eq:oks1}\\
    &\text{where }\evw_k =\frac{\evv_k}{\sum_{k=1}^K\evv_k}, \quad \text{and } \evl_k^2=\mathrm{var}_ka,\label{eq:oks} 
\end{align}
\noindent
\noindent where $a$ is the body area, $\mathrm{var}_k$ is a per-keypoint annotation falloff constant, and $\evv_k$ is a visibility indicator equal to 1 only if keypoint $k$ is present in the scene\footnote{Present includes both occluded and unoccluded keypoints within the bounding box crop.}.  The scaling $l_k$ within the exponential envelope accounts for differences in scale across the different body joints and overall area of the pose while the weighting $w_k$ takes presence in the scene into account.  

A person instance is regarded as correct (positive) if its OKS exceeds some threshold. Over a dataset with $N$ samples, we can tabulate the mean Average Recall (mAR) and mean Average Precision (mAP) as follows over $T$ thresholds $\{\tau_{t}\}$: 

\begin{equation}
    \mathrm{mAR} = \frac{1}{T}\sum_{t=1}^T\sum_{i=1}^N\frac{\mathcal{I}(c_i>\tau_{t})}N.
\end{equation}
\noindent
This equation clearly states that mAR is \textbf{ranking-independent} to the predicted confidence and purely evaluates the accuracy of poses. However, the primary metric used for evaluating 2D pose estimation is mean Average Precision (mAP).  The mAP is defined as
\begin{equation}
\mathrm{mAP}
    =\frac{1}{T}\sum_{t=1}^T\sum_{i'=1}^N\frac{\mathcal{I}(c_{i'}>\tau_t)}N\cdot\frac{\sum_{j=1}^{i'}{\mathcal{I}(c_j>\tau_t)}}{i'},\label{eq:map}
\end{equation}
where $i'$ denotes an index based on the instances sorted according to their estimated confidences, \ie $\hat{c}_1\geq\dots\hat{c}_{i'}\geq\dots\geq\hat{c}_N$.  The mAP therefore relies on the estimated confidences $\hat{c}$ to be consistent with the OKS in relative ordering and 
is \textbf{dependent} on the ranking of the predicted confidence.  
Note that this formulation of mAP is the same as the Area Under (maximum) Precision-Recall Curve (AUPRC) as in conventional classification~\citep{qi2021stochastic}\footnote{Note that another conventional metric AUROC=AR in the context.}.

\section{An Analysis on Pose Calibration} 
\label{sec:method}
\subsection{Motivation}
\label{sec:motivation}

As stated in \Eqref{eq:map}, a high mAP not only requires an accurate prediction but also a good alignment between confidence and OKS. Intuitively, to achieve the upper bound of mAP, the ranking of confidence should be the same of that of OKS. However, no attempts have been made in the pose estimation literature to systematically learn the impact of the confidence estimation. As a sanity test, we replace the estimated score in heatmap-based and RLE-based method with a constant value. As shown in the second and third columns of Table~\ref{tab:oks_form}, AP significantly drops around 5 point, which indicates the significant impact of confidence in the evaluation of AP.
With this observation, we start our method by theoretically understanding how well current methods predict the confidence compared to the OKS.

\subsection{Assumptions}

For the analysis, we formulate the expected OKS and expected confidence of both heatmap- and RLE-based methods from a statistical perspective. Specifically, we follow two standard assumptions~\citep{xiao2018simple,li2021human}. First, the $K$ keypoints of a person are conditionally independent, given the image itself. For clarity, we will drop the $k$ indices in this section. 
Secondly, we assume that the ground truth location of each keypoint in an image follows a Gaussian distribution 
$\rvp\sim\mathcal{N}(\vmu,\sigma^2\mI)$.
This Gaussian models ambiguities and errors in the annotation process~\citep{li2021human,chen2023mhentropy}, where $\vmu$ specifies the true underlying location.  For simplicity, we consider an isotropic Gaussian in our exposition and develop our analysis only in terms of variance $\sigma^2$, although the analysis can easily be extended for non-isotropic cases as well with a full covariance.

\subsection{Expected OKS}
We study the pose calibration problem by first deriving the expected OKS for a given estimated pose $\hat{\rvp}$ and the assumed Gaussian distribution for the ground truth pose $\rvp$. Based on the definition of OKS in \eqref{eq:oks1}, the expected OKS is given as 
\begin{align}
    \E_{\rvp}[\mathrm{OKS}]&=\E_{{\rvp}}\left[\exp\left(-\frac{\|\hat{\rvp}-\rvp\|^2}{2l^2}\right)\right]\\
    &=\frac{l^2}{\sigma^2+l^2}\exp\left(-\frac{\|\hat{\rvp}-\vmu\|^2}{2(\sigma^2+l^2)}\right),
    \label{eq:oks_mean}
\end{align}
\noindent
which is a function of  $\{\vmu,\sigma,l,\hat{\rvp}\}$. When a network is perfectly trained, $\hat{\rvp}$ will approach $\vmu$ and the corresponding OKS is represented as
\begin{equation}\label{eq:max_oks}
    s_{\text{OKS}} = \frac{l^2}{\sigma^2+l^2} = 1 - \frac{\sigma^2}{\sigma^2 + l^2}.
\end{equation}

\subsection{Expected Confidence}

\textbf{Heatmap} methods synthesize a ground truth heatmap {$\mH$} by constructing an isotropic Gaussian centered at the ground truth $\rvp$ and a standard deviation of $\tilde{l}$ set heuristically, \eg, $\tilde{l}\!=\!2$.
Given our previous assumption on the distribution of $\rvp$, the effective ground truth can be expressed as 
$\tilde{\ervp}\sim\mathcal{N}(\evmu,\evsigma^2+\tilde{l}^2)$ or in heatmap form as 
\begin{align}
    \rh_{\tilde{\rvp}}=2\pi \tilde{l}^2p(\tilde{\rvp}|\rvp)=\exp\left(-\frac{\|\tilde{\rvp}-\rvp\|^2}{2\tilde{l}^2}\right),
\end{align}

If we consider the predicted heatmap $\hat{h}$ as that which minimizes the MSE loss in \eqref{eq:detection_loss}, we arrive at the following: 
\begin{align}
    \hat{h}&=\argmin_{\hat{h}}\E_{\rvp}[(\hat{h}-\rh)^2]=\E_{\rvp}[\rh]\label{eq:likelihood}\\
    &=\int_{\rvp}p(\rvp|\rvx)\cdot2\pi \textbf{}^2p(\tilde{\rvp}|\rvp)\mathbf{d}\rvp=2\pi \textbf{}^2p(\tilde{\rvp}|\rvx).\label{eq:density}
\end{align}
The resulting optimal spatial heatmap $\hat{\mH}=\{\hat{h}\}\overset{c}{\approx}\mathcal{N}(\vmu,\hat{\sigma}^2\mI)$
with emerging $\hat{\sigma}^2=\sigma^2+\tilde{l}^2$, approximates rendered ground truth heatmap (see Supplementary for a similar derivation for the case with an imperfect mean).
This derivation highlights that predicted heatmaps learned with a pixel-wise MSE loss
exhibit a standard deviation slightly larger than $\tilde{l}=2$ even if the coordinate prediction is accurate~\citep{gu2021dive}. See Supp.~\tr{A} for a verification.

It follows from \eqref{eq:density} that the predicted confidence, defined as the max from~\eqref{eq:pre_heatmap} and is located at $\hat{\vp}\approx\vmu$ has a value given by,
\begin{align}
    \hat{\evs}_{\text{det}}&=\hat{\rh}_{\vmu}=2\pi \tilde{l}^2p(\tilde{\rvp}=\vmu|\rvx)\\
    &=\frac{2\pi \tilde{l}^2}{2\pi(\sigma^2+\tilde{l}^2)}\exp\left(-\frac{\|\vmu-\vmu\|^2}{2(\sigma^2+\tilde{l}^2)}\right)
    =\frac{\tilde{l}^2}{\sigma^2+\tilde{l}^2}=\frac{\tilde{l}^2}{\hat{\sigma}^2}.\label{eq:det_score}
\end{align}

The two expected values from~\eqref{eq:max_oks} and~\eqref{eq:det_score} are different at the same location $\vmu$. This difference comes from that $\tilde{l}$ is a constant value whereas $l$ changes according to different instance sizes and keypoints. 

\noindent\textbf{RLE-based regression} methods are learned by minimizing a negative log-likelihood over the predicted distribution as shown in~\eqref{eq:nll}. For simplicity, we {consider a normal distribution (an alternative distribution such as the Laplace or Normalizing Flow do not change the conclusions)}
$\rvp^{\prime}\sim$$\mathcal{N}(\hat{\vp},\hat{\sigma}^2\mI)$ are trained by  over distribution,
which is treated as a maximum optimization,
\begin{align}
\max_{\hat{\evp},\hat{\evsigma}}\E_{\rvp}\left[\log\frac1{\sqrt{2\pi}\hat{\evsigma}}\exp\left(-\frac{(\rvp-\hat{\rvp})^2}{2\hat{\evsigma}^2}\right)\right].\label{eq:rle_maximum}
\end{align}
When \eqref{eq:rle_maximum} is maximized, the predictive distribution approximates the optimal, $\hat{\evp}\approx\evmu,\hat{\evsigma}\approx\evsigma,\rvp^{\prime}\approx\rvp$. We give detailed derivations in the Supplementary.

Subsituting the $\hat{\sigma}$ from above into the heuristic score for RLE given in~\eqref{eq:pre_rle}, we arrive at
\begin{align}\label{eq:conf_rle}
    \hat{\evs}_{\text{reg}}=1-\hat{\sigma}=\E\left[1-\sqrt{\frac{\pi}8}\|\hat{\rvp}-\rvp\|_1\right].
\end{align}

Comparing~\eqref{eq:max_oks} with~\eqref{eq:conf_rle}, the expected confidence of RLE-based methods takes a linear form of $\hat{\sigma}$ and only models the annotation variation but ignores the instance size. 
Besides, it averages across all keypoints not excluding invisible ones, \ie, $\hat{\vv}=\mathbf{1}$, leading to more inconsistencies with OKS (\eqref{eq:oks}).

\subsection{Discussion}

\begin{table}[t]
  \centering
  \def\arraystretch{1.1}
  \scalebox{0.9}{
  \begin{tabular}{@{}lcccccc@{}}
    \toprule
    \multirow{2}{*}{\bf Method} & \multirow{2}{*}{\bf Orig}& \multirow{2}{*}{\bf \emph{const.}} & \multicolumn{3}{c}{\bf OKS} & \multirow{2}{*}{\bf mAR$\uparrow$} \\ \cline{4-6}
    & & &{\bf Mean} & {\bf Pred} & {\bf GT} & \\
    \midrule
    SBL~\citep{xiao2018simple} & 72.4 & 67.5 & 72.2 & 73.0 & 73.6 & 75.6 \\
    RLE~\citep{li2021human} & 72.2 & 67.2 & 71.8 & 73.2 & 73.3 & 75.4 \\
    \bottomrule
  \end{tabular}}
  \caption{mAP under different confidence estimation except the last column. 'Orig' means applying their original ways of estimating confidence. '\emph{consts.}' means replacing the confidence with a constant value. 'Mean', 'Pred', 'GT' corresponds to adjusting the confidence prediction with~\eqref{eq:max_oks} of mean area, predicted area (with additional supervision), and ground truth area. Results show that confidence estimation significantly influences the final evaluation and our closed-form adjustment can increase the mAP.}
  \label{tab:oks_form}
\end{table}

\paragraph{Misordered Ranking.} Although the three confidence forms all give descending curves, the different decay rates will influence the ranking. One explanation is that when it comes to a specific sample, the actual OKS will vary according to different samples but the expected confidences of both heatmap- and RLE-based methods remain unchanged since they don't consider the instance size and keypoint falloff constants. Image two similar $\sigma$s, OKS is likely to have different rankings depending on $l$, which becomes inconsistent with the ranking of expected confidences.

\paragraph{Confidence Correction.}

Motived by the above analysis, we provide a simple confidence correction to make the pose network better calibrated to the OKS based on the different confidence forms, which, at the same time, serves as the empirical verification of our theoretical derivations. With original $\hat{\sigma}$ (for detection 
 one way is to fit heatmap with Gaussian to get) and estimated or ground truth per sample $\evl$, scores can be adjusted to \eqref{eq:oks_mean}. Table~\ref{tab:oks_form} demonstrates that appropriately adjusting the confidence form to the ideal one improves mAP. Yet, this rescoring is based on the dismantling of metrics in ideal assumptions. We give the limitations as follows.

\begin{table*}[t]
  \centering
  \def\arraystretch{1.1}
  \resizebox{\textwidth}{!}{
  \begin{tabular}{lccccccccccc}
    \toprule
    {\bf Method} & {\bf Confidence} & {\bf Backbone} & {\bf Input Size} & {\bf\#Params (M)} & {\bf\#GFLOPs} & {\bf mAP$\uparrow$} & {\bf AP.5$\uparrow$} & {\bf AP.75$\uparrow$} & {\bf AP (M)$\uparrow$} & {\bf AP (L)$\uparrow$} & {\bf mAR$\uparrow$} \\
    \midrule
    \multicolumn{11}{l}{\bf Detection} \\ \hline
    SBL~\citep{xiao2018simple} & Hm & ResNet-50 & 256$\times$192 & 34.00 & 5.46 & 72.4& 91.5 & 80.4 & 69.8 & 76.6 &75.6\\
    \rowcolor{lightgray!50}{\bf+CCNet} &  & ResNet-50 & 256$\times$192 & 34.08 & 5.52 & {\color{blue}73.3 (+0.9)} & {\color{blue}92.6}& {\color{blue}80.9} & {\color{blue}70.4}& {\color{blue}77.5} &75.6\\
    SBL~\cite{xiao2018simple}&Hm & ResNet-152 & 384$\times$288 & 68.64 & 12.77 & 76.5  & 92.5 & 83.6 & 73.6 & 81.2 &79.3 \\
    \rowcolor{lightgray!50}{\bf+CCNet}& & ResNet-152 & 384$\times$288 & 68.71 & 12.83 & {\color{blue}77.3 (+0.8)}  & \tb{93.5} & \tb{84.1} & \tb{74.0} & \tb{81.6} &79.3\\
    HRNet~\cite{sun2019deep}&Hm & HRNet-W32 & 256$\times$192 & 28.54 & 7.7 & 76.0  & 93.5 & 83.4 & 73.7 & 80.0 & 79.3\\
    \rowcolor{lightgray!50}{\bf+CCNet}& & HRNet-W32 & 256$\times$192 & 28.62 & 7.76 & {\color{blue}77.0 (+1.0)}  & \tb{93.7} & \tb{84.0} & \tb{74.0} & \tb{81.0} & 79.3 \\
    HRNet~\cite{sun2019deep}&Hm & HRNet-W48 & 384$\times$288 & 63.62 & 15.31 & 77.4  & 93.4 & 84.4 & 74.8 & 82.1 & 80.9 \\
    \rowcolor{lightgray!50}{\bf+CCNet}& & HRNet-W48 & 384$\times$288 & 73.69 & 15.36 & {\color{blue}78.3 (+0.9)}  & \tb{93.6} & \tb{85.1} & \tb{75.5} & \tb{83.4}&80.9 \\
    ViTPose~\cite{xu2022vitpose} &Hm &ViT-Base & 256$\times$192& 89.99 & 17.85 & 77.3 & 93.5 & 84.5 & 75.0 & 81.6&80.4 \\
    \rowcolor{lightgray!50}{\bf+CCNet}& &ViT-Base & 256$\times$192& 90.07 & 17.91 & {\color{blue}78.1 (+0.8)} & \tb{93.7} & \tb{85.0} & \tb{75.4} & \tb{83.3}&80.4 \\
    \hline
    \multicolumn{11}{l}{\bf Regression} \\ \hline 
    {RLE~\citep{li2021human}} &Reg & {ResNet-50} & {256$\times$192} & 23.6 & 4.0 & 72.2 & 90.5 & 79.2 & 71.8 & 75.3 & {75.4} \\
    \rowcolor{lightgray!50}{\bf+CCNet}  & &{ResNet-50} &{256$\times$192} &23.6 &4.0 & {\color{blue}73.6 (+1.4)} & {\color{blue}91.6} & {\color{blue}80.2} & {\color{blue}72.0} & {\color{blue}77.6} & {75.4}  \\
    {RLE~\citep{li2021human}} & Reg & {ResNet-152} & {384$\times$288} & 58.3 & 11.3 & 76.3 & 92.4 & 82.6 & 75.6 & 79.7 & 79.2 \\
    \rowcolor{lightgray!50}{\bf+CCNet}  & &ResNet-152 &384$\times$288 & 58.3 & 11.3 & {\color{blue}77.1 (+0.8)} & {\color{blue}92.6} & {\color{blue}83.2} & 75.6 & {\color{blue}81.3} & 79.2 \\
    {RLE~\citep{li2021human}} & {Reg} & {HRNet-W32} & {256$\times$192} & 39.3 &7.1 & 76.7 & 92.4 & 83.5 & 76.0 & 79.3 & {79.4} \\
    \rowcolor{lightgray!50}{\bf+CCNet}& &{HRNet-W32} & {256$\times$192} & 39.3 & 7.1& {\color{blue}77.5 (+0.8)} & {\color{blue}92.6} & {\color{blue}84.2} & {75.9} & {\color{blue}81.3} & {79.4}  \\
    {RLE~\citep{li2021human}} &Reg & {HRNet-W48} & {384$\times$288} & 75.6 & 33.3 & 77.9 & 92.4 & 84.5 & 77.1 & 81.4 & {80.6} \\
    \rowcolor{lightgray!50}{\bf+CCNet}& &{HRNet-W48} & {384$\times$288}& 75.6 & 33.3 & {\color{blue}78.8 (+0.9)} & {\color{blue}92.6} & {\color{blue}85.1} & {77.0} & {\color{blue}82.9} & {80.6} \\
    {Poseur~\citep{mao2022poseur}} &Reg & {ResNet-50} & {256$\times$192} &33.1 & 4.6 & 76.8 & 92.6 & 83.7 & 74.2 & 81.4 & {79.7} \\
    \rowcolor{lightgray!50}{\bf+CCNet}&&{ResNet-50} & {256$\times$192}& 33.1 & 4.6& {\color{blue}77.7 (+0.9)} & {\color{blue}92.7} & {\color{blue}84.2} & {\color{blue}74.9} & {\color{blue}82.3} & {79.7} \\ \hline
    IPR~\citep{sun2018integral} & \emph{const.} &  ResNet-50 &{256$\times$192}& 34.0 & 5.5 & 65.6  & 88.1  & 71.8 & 61.3 & 70.2&74.9 \\
    IPR~\citep{sun2018integral} & \multirow{1}{*}{Hm} & ResNet-50 &{256$\times$192}& 34.0 & 5.5 & 69.5  & 88.9 & 74.6 & 67.2 & 74.7&74.9 \\ 
    \rowcolor{lightgray!50}{\bf+CCNet} &  &ResNet-50 &{256$\times$192}& 34.1 & 5.5 & \tb{70.8 (+1.3)}  & \tb{90.5} & \tb{78.1} & \tb{68.1} & \tb{75.8} & 74.9 \\ 
    \bottomrule
  \end{tabular}}
  \caption{Comparisons with state-of-the-art methods on the COCO validation set. The {\color{blue}blue} color depicts improved value after applying the proposed CCNet. ``Hm'', ``Reg'', and ''\emph{const.}'' represent confidence functions originating from heatmap maximum, direct regression, and constant value, respectively. This table demonstrates that our CCNet considerably improves the AP of all methods.}
  \label{tab:sotas}
\end{table*}

\section{ConfidenceNet}

The above analysis shows that changing the score form will improve the ranking and mAP. However, it is not sufficient because the analysis is conditioned on a perfect network that will achieve the optimum value w.r.t. the optimization objective, whereas in practice, different models present different correlations between prediction and $\hat{\sigma}$, which needs extra modeling.

Motivated by this, we propose a calibrated ConfidenceNet (CCNet), which designs a general, efficient, and effective calibration branch for existing pose estimation methods. Specifically, denoting the previous pose network as PredNet $\mathcal{P}$ which outputs keypoint locations, we apply a lightweight network ConfidenceNet $\mathcal{C}$ to predict confidence based on the features in $\mathcal{P}$. For instance, for heatmap-based methods, we detach and utilize the penultimate features after deconvolution layers; for RLE-based methods, we similarly use the features after the Global Average Pooling layer. In this way, it does not require re-training and allows ConfidenceNet $\mathcal{C}$ to get access to the rich features from the PredNet $\mathcal{P}$. Also, the PredNet $\mathcal{P}$ is fixed so mAR will not be affected.

Formally, the ConfidenceNet $\mathcal{C}$ outputs a calibrated confidence $\hat{\evs}_k\in[0,1]$ for each keypoint. Apart from keypoint confidence, it also predicts keypoint visibility $\hat{\evv}_k\in[0,1]$. This corrects the bias caused by thresholding confidence as visibility in existing practice (\eqref{eq:thres_vis_weight}). We observe that accuracy may not be well aligned with visibility (Sec.~\ref{sec:design_choices}).
For confidence, a simple yet effective MSE loss is applied to calibrate predictions with ground truth keypoint as,
\begin{align}
    \mathcal{L}_{\text{conf}}=\sum_{k=1}^K(\hat{\evs}_k-\evs_k)^2.\label{eq:oks_cal}
\end{align}
\noindent
where $\evs_k$ is the OKS for this keypoint.
The experiments find that the loss choices of calibration are robust to others including aleatory and Cross-Entropy. For visibility, we commonly treat it as a binary classification to use a Binary Cross-Entropy (BCE) loss as,
\begin{align}
    \mathcal{L}_{\text{vis}}=-\sum_{k=}^K(\evv_k\log\hat{\evv}_k+(1-\evv_k)\log(1-\hat{\evv}_k)).
\end{align}
The total loss is then a weighted sum as
\begin{align}
\mathcal{L}=\mathcal{L}_{\text{conf}}+\lambda\mathcal{L}_{\text{vis}}.
\end{align}
Following the OKS form (\eqref{eq:oks}), we similarly obtain the instance-level confidence by aggregating the predicted visibility and confidence.

\section{Experiments}\label{sec:experiments}

\subsection{Datasets}

\paragraph{Datasets \& Evaluation Metrics.} We evaluate the pose estimation task on three human pose benchmarks, MSCOCO~\cite{lin2014microsoft}, MPII~\cite{andriluka20142d}, MSCOCO-WholeBody~\cite{jin2020whole}. For the downstream tasks, we evaluate
the 3D fitting task on 3DPW.

The \textbf{MSCOCO} dataset consists of 250k person instances annotated with 17 keypoints. We follow the standard and evaluate the model with mAP over 10 OKS thresholds. To demonstrate the generalization ability of our method, we also evaluate our method on the \textbf{MPII} dataset with {Percentage of Correct Keypoints (PCK)} and the \textbf{MSCOCO-WholeBody} dataset includes face and hand annotations for each human instance, we test our method with the common metric mAP to show its capability on face and hand keypoint detection apart from the body.

For the downstream benchmarks,
\textbf{3DPW} is a more challenging outdoor benchmark, which provides around 3k SMPL annotations for testing. We follow the convention~\cite{kolotouros2019learning} and use MPJPE, PA-MPJPE, and MVE to measure the quality of the predicted 3D mesh.  Additional implementation details and pseudo-code is provided in the Supplementary.

\subsection{Comparions with SOTAs}

\paragraph{Evaluation on MSCOCO~\citep{lin2014microsoft}.} Since our method is a plug-and-play module after the training of pose models, we evaluate our method on several baselines, including SBL~\cite{xiao2018simple}, HRNet~\cite{sun2019deep}, VitPose~\cite{xu2022vitpose} for heatmap-based pipelines, RLE~\cite{li2021human}, IPR~\cite{sun2018integral}, Poseur~\cite{mao2022poseur} for regression-based pipelines. We perform our method based on the checkpoints released on the official website. We report our results in~\cref{tab:sotas}. Our simple yet effective method universally improves across varying backbones, learning pipelines, and scoring functions, which means it can be applied model-agnostically even when the uncertainty estimation capabilities of different networks may vary from PredNets to PredNets. Also, we argue that pose estimation methods should be aware of confidence estimation and report both improved mAP and mAR. The gap between the two reasonably shows how calibrated the pose model is.

\begin{table}[t]
  \centering
  \def\arraystretch{1.1}
  \resizebox{0.95\linewidth}{!}{
  \begin{tabular}{llccccc}
    \toprule
    & & {\bf Body} & {\bf Foot} & {\bf Face} & {\bf Hand} & {\bf Whole} \\
    \midrule
    \multirow{3}{*}{\bf mAP$\uparrow$} & Whole~\citep{mao2022poseur} & 67.2 & 63.6 & 84.6 & 58.3 & 61.0 \\
    & Part & 68.5 & 68.9 & 85.9 & 62.5 & 61.0 \\
    \rowcolor{lightgray!50} & {\bf+CCNet} & {\color{blue}69.9} & {\color{blue}69.2} & {\color{blue}86.4} & {\color{blue}62.7} & {\color{blue}63.3 (+2.3)} \\
    {\bf mAR$\uparrow$} & & 72.3 & 72.9 & 88.1 & 65.4 & 67.2 \\
    \bottomrule
  \end{tabular}}
  \caption{mAP evaluation on the COCO-WholeBody validation set based on Poseur~\citep{mao2022poseur}. We base our CCNet on the part confidence and improve the whole body AP by 2.3.}
  \label{tab:cocowb}
\end{table}

\begin{table}[t]
  \centering
  \def\arraystretch{1.1}
  \resizebox{0.95\linewidth}{!}{
  \begin{tabular}{lcccccc}
    \toprule
    & \multirow{2}{*}{\bf PCK.5$\uparrow$} & \multirow{2}{*}{\bf PCK.1$\uparrow$} & \multirow{2}{*}{\bf mAP$\uparrow$} & \multirow{2}{*}{\bf mAR$\uparrow$} & \multicolumn{2}{c}{\bf AUSE$\downarrow$} \\ \cline{6-7}
    & & & & & {\bf PCK.5} & {\bf PCK.1} \\
    \midrule
    RLE~\citep{li2021human} & {86.2} & {32.9} & 75.4 & {78.8} & 3.35 & 1.76 \\
    \rowcolor{lightgray!50} {\bf+CCNet} & & & {\color{blue}76.6 (+1.2)} & & {\color{blue}2.98} & {\color{blue}1.49} \\
    SBL~\citep{li2021human} & {88.5} & {33.9} & 77.3 & {80.5} & 3.90 & 2.36 \\
    \rowcolor{lightgray!50} {\bf+CCNet} & & & {\color{blue}77.7 (+0.4)} & & {\color{blue}3.52} & {\color{blue}1.95} \\
    \bottomrule
  \end{tabular}}
  \caption{The proposed CCNet improves all mAP and AUSE-PCK evaluations on the MPII validation set.}
  \label{tab:mpii}
\end{table}

\begin{table}[t]
  \centering
  \def\arraystretch{1.1}
  \resizebox{0.95\linewidth}{!}{
  \begin{tabular}{lcccccc}
    \toprule
    & {\bf mAP$\uparrow$} & {\bf mAR$\uparrow$} & \multicolumn{2}{c}{\bf Pearson Corr$\uparrow$} & \multicolumn{2}{c}{\bf AUSE-OKS$\downarrow$} \\ \cline{2-3}\cline{4-5}\cline{6-7}
    & \multicolumn{2}{c}{\bf KP} & {\bf Ins} & {\bf KP} & {\bf Ins} & {\bf KP} \\
    \midrule
    RLE~\citep{li2021human} & 77.9 & 82.7 & 0.700 & 0.637 & 2.72 & 5.03 \\
    \rowcolor{lightgray!50}{\bf+CCNet} & {\color{blue}78.7 (+0.8)} & & {\color{blue}0.782} & 0.636 & {\color{blue}1.72} & {\color{blue}4.22} \\
    SBL~\citep{xiao2018simple} & 76.7 & 82.8 & 0.643 & 0.543 & 2.77 & 6.47 \\
    \rowcolor{lightgray!50}{\bf+CCNet} & {\color{blue}78.9 (+2.2)} & & {\color{blue}0.718} & {\color{blue}0.628} & {\color{blue}2.13} & {\color{blue}4.11} \\
    \bottomrule
  \end{tabular}}
  \caption{Other confidence quantification evaluations except mAP on the COCO validation set, where ``Ins'' and ``KP'' are short for instance and keypoint, respectively.}
  \label{tab:other_metrics}
\end{table}

\paragraph{COCO-WholeBody~\citep{jin2020whole}} dataset evaluates the task of whole-body pose estimation, which includes not only body keypoints but also face and hand keypoints. The confidence estimation is even less considered. For example, the convention is to assign the whole instance confidence to each part, which is not reasonable when evaluating the AP of the corresponding part. By simply changing the confidence of each part to the aggregation of the predicted part (the third row in~\cref{tab:cocowb}) instead of all keypoints (the second row in~\cref{tab:cocowb}), the AP is significantly improved up to 5.3 for foot. Applying the proposed CCNet, we further improve the AP on every part and the whole body.

\paragraph{MPII~\citep{andriluka20142d}} is a single-person dataset and another commonly used benchmark (Tab.~\ref{tab:mpii}). OKS is defined similarly on MPII, where the tight boundary box is used as the area~\citep{jin2020whole,lin2014microsoft}. Though PCK~\citep{andriluka20142d}, average of visible $\mathcal{I}(\|\vr_k\|\leq\tau)$, as a simpler metric is less comprehensive than OKS~\citep{lin2014microsoft,ruggero2017benchmarking,shi2022end}, we also try to calibrate the model \wrt it to verify the applicability of our method to other metrics and datasets. Since no mAP is defined on PCK, we evaluate with another metric AUSE~\citep{ilg2018uncertainty}.

\begin{figure}[t]
    \centering
    \includegraphics[width=0.95\linewidth]{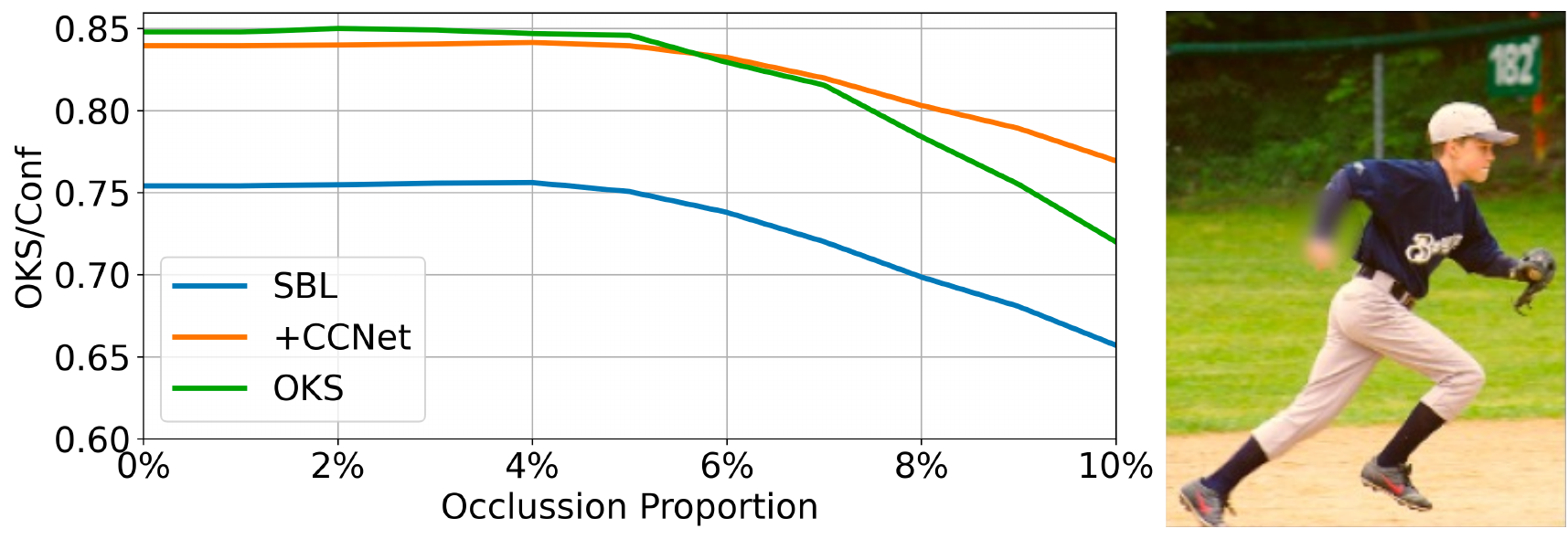}
    \caption{OKS and estimated confidence decrease are well correlated with the occluder size proportional to the input size. The right panel gives an illustration of a Gaussian blur placed on the right wrist.}
    \label{fig:occ}
\end{figure}

\paragraph{Other Confidence Testings.} Apart from the most established mAP, (1) Pearson Correlation testing between instance/keypoint OKS and their confidence is also adopted to evaluate the reliability of confidence~\citep{li2021human,gu2021dive,bramlage2023plausible}. (2) We measure Area Under Sparsification Error (AUSE)~\citep{ilg2018uncertainty,franchi2022muad} plotted by gradually removing the most uncertain samples and computing the remaining error ($1-\text{OKS}$). (3) As the occlusion patch expands, confidence is expected to shrink along with OKS (Fig.~\ref{fig:occ}). The outperformance of our method across all tests as expected shows confidence is really calibrated (Tab.~\ref{tab:other_metrics}).
Quantitative visualization of calibrated confidence is provided in the Supp. ~\tr{C}.

\subsection{Confidence Estimation for Downstream Tasks}

Better uncertainty estimation and calibration benefit downstream tasks such as 3D model fitting. We explore the application of confidence calibration in pose estimation.

\begin{table}[t]
  \centering
  \def\arraystretch{1.1}
  \resizebox{0.7\linewidth}{!}{
  \begin{tabular}{lccc}
    \toprule
    {\bf Method} & {\bf PA-MPJPE$\downarrow$} & {\bf MPJPE$\downarrow$} & {\bf MVE$\downarrow$} \\
    \midrule
    SPIN~\cite{kolotouros2019learning} & 60.2 & 102.1 & 130.6 \\
    +SBL~\cite{xiao2018simple} & 58.8 & 100.5 & 128.7 \\
    \rowcolor{lightgray!50} {\bf+CCNet} & {\color{blue}57.8} & {\color{blue}99.7} & {\color{blue}127.5} \\
    \bottomrule
  \end{tabular}}
  \caption{3D errors on 3DPW test set. Results show that better calibrated 2D pose net further improves the 3D results. }
  \label{tab:model_fitting}
  \vspace{-.5cm}
\end{table}

\paragraph{3D Mesh Fitting.} Mesh recovery tasks often employ 2D keypoints to optimize the output 3D pose and shape. Given 2D detection results from the off-the-shelf pose network~\cite{xiao2018simple}, the 2D reprojection loss usually consists of a weighted sum between the reprojected 2D predictions from the 3D mesh and its corresponding 2D detection results where the weight is exactly the predicted confidence. We follow CLIFF to refine the initial predictions from SPIN. As shown in (Tab.~\ref{tab:model_fitting}), the calibrated 2D pose network better refines the 3D predictions.

\subsection{Design Choices \& Discussions}\label{sec:design_choices}

\begin{figure}[t]
    \centering
    \begin{overpic}[width=\linewidth]{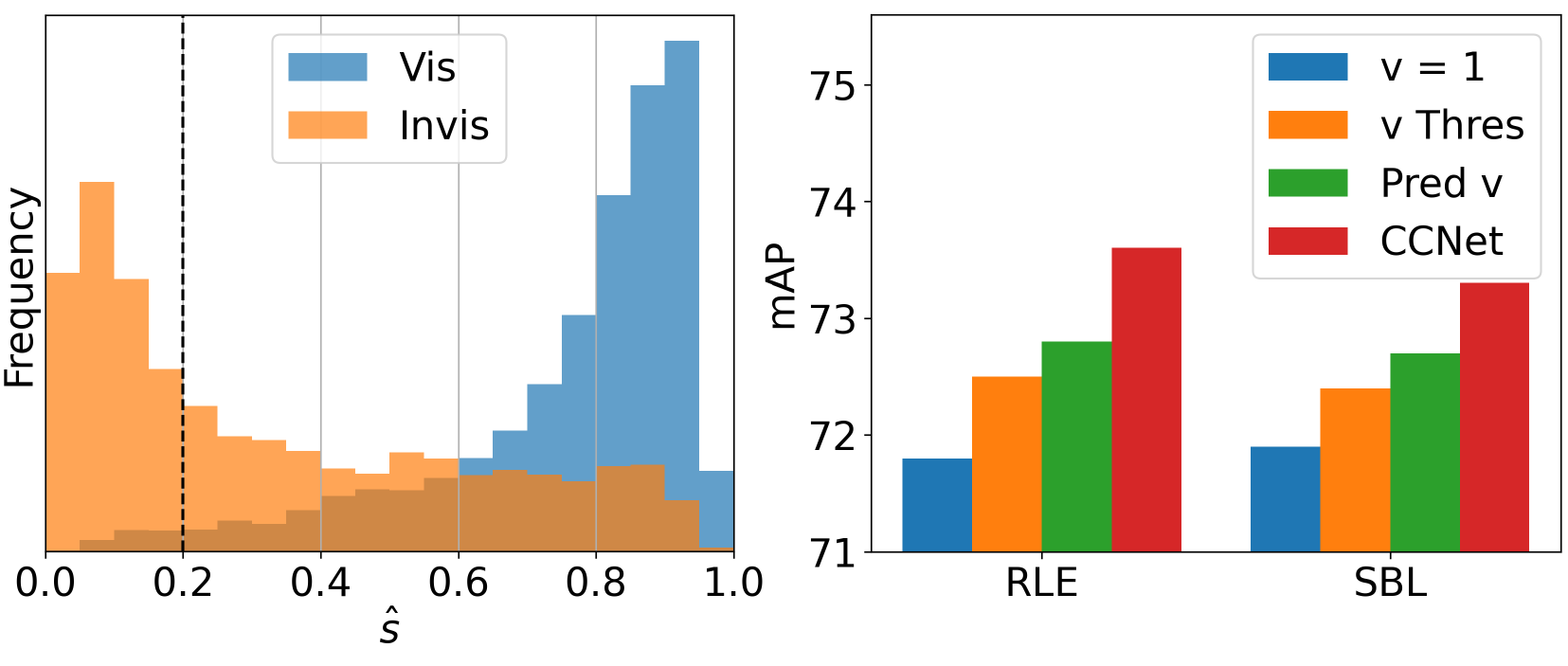}
        \put(23,-2){\small\bf (a)\label{fig:oks_vis}}
        \put(76,-2){\small\bf (b)\label{fig:aggregation}}
    \end{overpic}
    \caption{Visibility aggregation ablation. \textbf{(a)} Confidence distribution of visible and invisible keypoints. \textbf{(b)} Aggregations from different visibilities verify the effectiveness of visibility prediction in keypoint confidence aggregation. }
    \label{fig:agg_ablation}
\end{figure}

\paragraph{Surrogate Losses}~\cite{kendall2017uncertainties,lakshminarayanan2017simple,corbiere2019addressing,amini2020deep,qi2021stochastic,yu2021slurp} were proposed to tackle the confidence estimation task. In our exploration, surprisingly, we find these sophisticated methods capture uncertainty no better than MSE. This might be bound by post-hoc confidence estimation of a given PredNet and estimatability~\cite{yu2023discretization}. Note that pre-training confidence estimation is generally observed to hurt prediction accuracy as well as lead to a less satisfactory mAP~\cite{bramlage2023plausible,pathiraja2023multiclass}. How to better hybridize the advantages of these two leaves a promising future work.

\paragraph{Input Features.} Generally, confidence estimation is based on penultimate features. This promotes the lightweight of the CCNet. To verify that they contain sufficient rich information, we did the comparisons with additional input low-level features of shallow layers, prediction and original keypoint confidence roughly indicating the groundtruth range. A strategy of copying the backbone and fine-tuning similar to \citet{corbiere2019addressing,yu2021slurp,zhang2023adding} is also considered. As a result, we find that penultimate feature input is sufficient~\citep{yu2023discretization}.

\paragraph{Confidence Aggregation.} Existing work~\cite{xiao2018simple,gu2021dive} empirically set STD of rendered Gaussian heatmaps and a visibility threshold based on confidence estimate. They found the network is sensitive to the choice of the hyperparameters. Furthermore, the model has a different inductive bias from the human; keypoints with low confidence are not necessarily invisible (Fig.~\ref{fig:agg_ablation} (a)). Thus, different common aggregations
are studied in Fig.~\ref{fig:agg_ablation}(b). The visibility classification strategy shows promising performance without much burden brought. Additional OKS calibration (\cref{eq:oks_cal}) further pushes up the performance.

\section{Conclusion}\label{sec:discussions}

In this paper, we are the first to tackle the pose calibration problem which requires the predicted confidence to be aligned with the accuracy metric OKS. By deriving the expected value of OKS under a general assumption, we theoretically reveal that heuristic ways of getting the confidence in current pose estimation methods cause a misalignment. Although this misalignment can be alleviated by predicting the instance size and adjusting the confidence form, it is not sufficient due to black-box nature of neural networks. Therefore, we propose a Calibrated ConfidenceNet (CCNet) to address all the issues and learn a network-aware branch to align the OKS. Our experiments demonstrate that CCNet is applicable to all pose methods on various datasets and the better calibrated confidence will also help downstream outputs like 3D keypoints.

\newpage
{
    \small
    \bibliographystyle{ieeenat_fullname}
    \bibliography{main}
}

\clearpage
\setcounter{page}{1}
\maketitlesupplementary

\setcounter{section}{0}
\renewcommand{\thesection}{\Alph{section}}
\renewcommand{\thefigure}{\alph{figure}}
\renewcommand{\thetable}{\alph{table}}

The supplementary materials include \textbf{A.} Theoretical Understanding, \textbf{B.} Implementation Details, and \textbf{C.} More Experimental Results, referred in the manuscript.

\section{Theoretical Understanding}

\subsection{Illustration of Setting}
\begin{figure}[t]
    \centering
    \includegraphics[width=\linewidth]{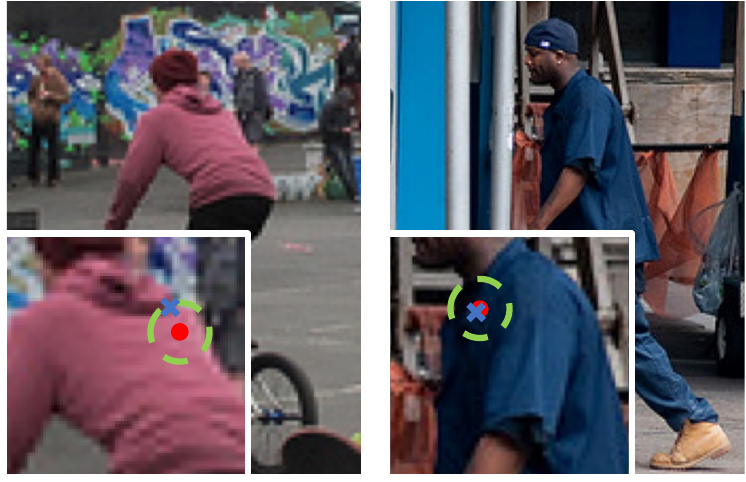}
    \caption{An illustration of visual cues that have similar (underlying/predictive) uncertainty ({\color{red} red circle} mean and {\color{green} green dashed circle} range) but different per-sample annotation ({\color{blue} blue} crosses) treated as a sample from the distribution. For instance, the left one is further from the mean than the right one.}
    \label{fig:ill}
\end{figure}

Different from 1 image $\rvx$ corresponding to only 1 pose keypoint $\rvp$, we consider stochastics caused by annotation error and occlusion ambiguity, \etc, by a 1-to-many distribution $p(\rvp|\rvx)$ (L275-276). Specifically, for two inputs $\vx_1,\vx_2$ with similar ambiguity $\sigma_1\approx\sigma_2$, accuracy (\eg, OKS) may be different sample-wisely (\Cref{fig:ill}), but they are supposed to have similar rankings regardless of uncontrollable and irreducible uncertainty~\citep{kendall2017uncertainties}. Think about it from another perspective: if the person is asked to re-annotate the two images, analogous to \emph{re-sampling} of the distribution, the accuracy of the first image has chance of being higher than that of the second. The goal is to achieve the highest mAP in the expected sense of distributions. 

\subsection{Expected OKS \Cref{eq:oks_mean}}
\begin{proof}
It follows a Normal distribution (L321-323); the integral is also tractable to compute as shown below.
\begin{align}
    &\E_{\rvp\sim\mathcal{N}(\boldsymbol{\mu},\sigma^2\mI)}[\text{OKS}]\\
    =&\int_{\rvp}\frac1{2\pi\sigma^2}\exp\left(-\frac{\|\rvp-\boldsymbol{\mu}\|^2}{2\sigma^2}\right)\exp\left(-\frac{\|\rvp-\hat{\rvp}\|^2}{2\ervl^2}\right)\mathbf{d}\rvp\\
    =&\frac1{2\pi\sigma^2}\int_{\rvp}\exp\left(-\frac{\|\rvp-\boldsymbol{\mu}\|^2}{2\sigma^2}-\frac{\|\rvp-\hat{\rvp}\|^2}{2\ervl^2}\right)\mathbf{d}\rvp.\label{eq:oks_int_raw}
\end{align}

\begin{lemma}\label{lem:pphat}
In L300 of manuscript, the form is regarded as resemblingly the random variable $\hat{\rvp}\sim\mathcal{N}(\bm{\mu},({\sigma}^2+\ervl^2)\mI)$. \Ie,

\begin{align}
    \int_{\rvp}\mathcal{N}(\rvp|\bm{\mu},\sigma^2\mI)\mathcal{N}(\hat{\rvp}|\rvp,\ervl^2\mI)\mathbf{d}\rvp=\mathcal{N}(\hat{\rvp}|\bm{\mu},(\sigma^2+\ervl^2)\mI)
\end{align}
\begin{align}
    \iff&\int_{\rvp}\exp\left(-\frac{\|\rvp-\bm{\mu}\|^2}{2\sigma^2}\right)\exp\left(-\frac{\|\hat{\rvp}-\rvp\|^2}{2\ervl^2}\right)\mathbf{d}\rvp\\
    =&\frac{2\pi\sigma^2\ervl^2}{(\sigma^2+\ervl^2)}\exp\left(-\frac{\|\hat{\rvp}-\bm{\mu}\|^2}{2(\sigma^2+\ervl^2)}\right).
\end{align}
\end{lemma}

\begin{lemma}\label{lem:pp}
In another perspective, term within $\exp$ of \Cref{eq:oks_int_raw} can be also arranged \wrt $\rvp$ as
\begin{align}
    -\|\mathcal{D}\rvp-\overrightarrow{\mathcal{E}}\|^2-\mathcal{F},\label{eq:exp_arrange}
\end{align}
\begin{align}
    \mathcal{D}=\frac1{\sqrt{2\frac{\sigma^2\ervl^2}{\sigma^2+\ervl^2}}},
    \overrightarrow{\mathcal{E}}=\frac{\ervl^2\boldsymbol{\mu}+\sigma^2\hat{\rvp}}{\sqrt{2(\ervl^2+\sigma^2)\ervl^2\sigma^2}},
    \mathcal{F}=\frac{\|\hat{\rvp}-\boldsymbol{\mu}\|^2}{2(\sigma^2+\ervl^2)}.
\end{align}
\end{lemma}

Substituting back into \Cref{eq:oks_int_raw} obtains
\begin{align}
    &\frac1{2\pi\sigma^2}\int_{\rvp}\exp(-\|\mathcal{D}\rvp-\overrightarrow{\mathcal{E}}\|^2-\mathcal{F})\mathbf{d}\rvp\\
    =&\frac1{2\pi\sigma^2}\int_{\mathcal{D}\rvp-\overrightarrow{\mathcal{E}}}\exp(-\|\mathcal{D}\rvp-\overrightarrow{\mathcal{E}}\|^2)\exp(-\mathcal{F})\frac1{\mathcal{D}}\mathbf{d}\mathcal{D}\rvp-\overrightarrow{\mathcal{E}}\\
    =&\frac{\exp(-\mathcal{F})}{2\pi\sigma^2\mathcal{D}}\int_{\mathcal{D}\rvp-\overrightarrow{\mathcal{E}}}\exp(-\|\mathcal{D}\rvp-\overrightarrow{\mathcal{E}}\|^2)\mathbf{d}\mathcal{D}\rvp-\overrightarrow{\mathcal{E}}\label{eq:df_const2p}\\
    =&\frac{\exp(-\mathcal{F})}{2\pi\sigma^2\mathcal{D}}2\pi\frac12\\
    =&\frac{\ervl^2}{\sigma^2+\ervl^2}\exp\left(-\frac{\|\hat{\rvp}-\boldsymbol{\mu}\|^2}{2(\sigma^2+\ervl^2)}\right),\label{eq:sigmasqrt0.5}
\end{align}
where $\mathcal{D},\mathcal{F}$ are independent of $\rvp$ conditional on the image (\Cref{eq:df_const2p}); \Eqref{eq:sigmasqrt0.5} is based on
\begin{align}
    \int_{\rvx}\frac1{2\pi\frac12}\exp(-\|\rvx\|^2)\mathbf{d}\rvx=1.
\end{align}
\end{proof}

\subsection{Verification of Detection \texorpdfstring{$\hat{\sigma}^2=\sigma^2+\tilde{l}^2$}{} (L308)}
\begin{figure}[t]
    \centering
    \includegraphics[width=0.9\linewidth]{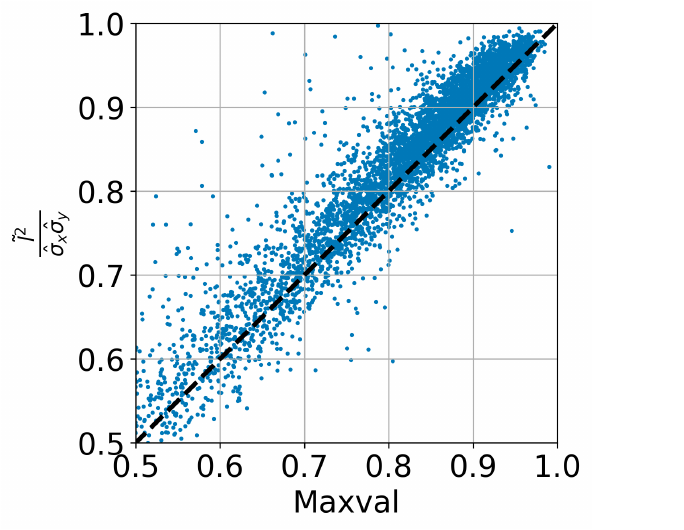}
    \caption{Maximum values of the heatmap are almost coincident with our estimated scoring (peak density as \Cref{eq:det_score}), which verifies derivation.}
    \label{fig:maxval_est}
\end{figure}

Figure~\ref{fig:maxval_est} verifies \Cref{eq:det_score} and model distribution (or heatmap) approximates noisy ground truth distribution instead of the pure one. Following~\cite{wehrbein2021probabilistic}, sigmas are estimated by fitting heatmap with Gaussian.

\subsection{Optima of \Cref{eq:rle_maximum} NLL}
\begin{proof}
It is well-established, but we still include it here for the convenience of readers. Formally,
\begin{align}
    \hat{\rvp}^{\ast},\hat{\sigma}^{\ast}=\argmin_{\hat{\rvp},\hat{\sigma}}\mathcal{L}_{\rm{nll}}=\argmax_{\hat{\rvp},\hat{\sigma}}\mathcal{L}_{\rm{ll}},
\end{align}
where in more general 2D case (1D in the manuscript for illustration), Log-Likelihood
\begin{align}
    \mathcal{L}_{\rm{ll}}&=\E_{\rvp\sim\mathcal{N}(\boldsymbol{\mu},\sigma^2\mI)}\left[\log\frac1{2\pi\hat{\sigma}^2}\exp\left(-\frac{\|\rvp-\hat{\rvp}\|^2}{2\hat{\sigma}^2}\right)\right]\\
    &=\E_{\rvp}\left[-\log2\pi-\log\hat{\sigma}^2-\frac{\|\rvp-\hat{\rvp}\|^2}{2\hat{\sigma}^2}\right]\\
    &\overset{c}{=}-\E_{\rvp}\left[\log\hat{\sigma}^2+\frac{\|\rvp-\hat{\rvp}\|^2}{2\hat{\sigma}^2}\right].
\end{align}

Denote
\begin{align}
    \mathcal{A}=\mathcal{B}+\mathcal{C},\mathcal{B}=\log\hat{\sigma}^2,\mathcal{C}=\frac{\|\rvp-\hat{\rvp}\|^2}{2\hat{\sigma}^2}.
\end{align}
The following is calculated:
\begin{align}
    \frac{\partial\mathcal{B}}{\partial\hat{\rvp}}=\mathbf{0},\frac{\partial\mathcal{B}}{\partial\hat{\sigma}^2}=\frac1{\hat{\sigma}^2},
\end{align}
\begin{align}
    \frac{\partial\mathcal{C}}{\partial\hat{\rvp}}&=\frac1{2\hat{\sigma}^2}\frac{\partial\|\rvp-\hat{\rvp}\|^2}{\partial\hat{\rvp}}=\frac1{2\hat{\sigma}^2}\frac{\partial\|\rvp-\hat{\rvp}\|^2}{\partial\rvp-\hat{\rvp}}\frac{\partial\rvp-\hat{\rvp}}{\partial\hat{\rvp}}\\
    &=\frac1{2\hat{\sigma}^2}2(\rvp-\hat{\rvp})^T(-\mI)=-\frac{\rvp-\hat{\rvp}}{\hat{\sigma}^2},
\end{align}
\begin{align}
    \frac{\partial\mathcal{C}}{\partial\hat{\sigma}^2}&=\frac{\|\rvp-\hat{\rvp}\|^2}2\frac{\partial\frac1{\hat{\sigma}^2}}{\partial\hat{\sigma}^2}=\frac{\|\rvp-\hat{\rvp}\|^2}2\left(-\frac1{\hat{\sigma}^4}\right)\\
    &=-\frac{\|\rvp-\hat{\rvp}\|^2}{2\hat{\sigma}^4}.\label{eq:pc_psigmahat}
\end{align}

\paragraph{Optimal $\hat{\rvp}$.} Taking derivative of $\mathcal{L}_{\rm{ll}}$ \wrt $\hat{\rvp}$ and setting it to 0 give
\begin{align}
    \frac{\partial\mathcal{L}_{\rm{ll}}}{\partial\hat{\rvp}}&=\frac{\partial-\E_{\rvp}\left[\mathcal{A}\right]}{\partial\hat{\rvp}}
    =-\E_{\rvp}\left[\frac{\partial\mathcal{A}}{\partial\hat{\rvp}}\right]\label{eq:p_const2phat}\\
    &=-\E_{\rvp}\left[\frac{\partial\mathcal{B}}{\partial\hat{\rvp}}+\frac{\partial\mathcal{C}}{\partial\hat{\rvp}}\right]=-\E_{\rvp}\left[\mathbf{0}-\frac{\rvp-\hat{\rvp}}{\hat{\sigma}^2}\right]\\
    &=\E_{\rvp}\left[\frac{\rvp-\hat{\rvp}}{\hat{\sigma}^2}\right]=\frac1{\hat{\sigma}^2}(\E_{\rvp}[\rvp]-\hat{\rvp})=\frac1{\hat{\sigma}^2}(\boldsymbol{\mu}-\hat{\rvp})\label{eq:phat_sigmahat_const2p}\\
    &=\mathbf{0}.
\end{align}
The facts that given an image, $\rvp$ in expectation is constant \wrt $\hat{\rvp}$ and $\hat{\rvp},\hat{\sigma}^2$ are constant \wrt $\rvp$ are used in \Cref{eq:p_const2phat,eq:phat_sigmahat_const2p}, respectively.
Thus, rearrangement gives optima
\begin{align}
    \hat{\rvp}^{\ast}=\boldsymbol{\mu}.\label{eq:opt_phat}
\end{align}

\paragraph{Optimal $\hat{\sigma}$.}Similarly, we derive derivative of $\mathcal{L}_{\rm{ll}}$ \wrt $\hat{\sigma}^2$ as
\begin{align}
    \frac{\partial\mathcal{L}_{\rm{ll}}}{\partial\hat{\sigma}^2}&=\frac{\partial-\E_{\rvp}\left[\mathcal{A}\right]}{\partial\hat{\sigma}^2}
    =-\E_{\rvp}\left[\frac{\partial\mathcal{A}}{\partial\hat{\sigma}^2}\right]\\
    &=-\E_{\rvp}\left[\frac{\partial\mathcal{B}}{\partial\hat{\sigma}^2}+\frac{\partial\mathcal{C}}{\partial\hat{\sigma}^2}\right]=-\E_{\rvp}\left[\frac1{\hat{\sigma}^2}-\frac{\|\rvp-\hat{\rvp}\|^2}{2\hat{\sigma}^4}\right]\\
    &=-\frac1{\hat{\sigma}^2}+\frac1{2\hat{\sigma}^4}\E_{\rvp}[\|\rvp-\hat{\rvp}\|^2].\label{eq:opt_sigma}
\end{align}
\Eqref{eq:opt_phat} optimal $\hat{\rvp}^{\ast}$ helps simplify it as
\begin{align}
    &-\frac1{\hat{\sigma}^2}+\frac1{2\hat{\sigma}^4}\E_{\rvp}[\|\rvp-\boldsymbol{\mu}\|^2]=-\frac1{\hat{\sigma}^2}+\frac1{2\hat{\sigma}^4}2\sigma^2\label{eq:var}\\
    &=-\frac1{\hat{\sigma}^2}+\frac{\sigma^2}{\hat{\sigma}^4}.
\end{align}
For \Cref{eq:var} the variance of the Normal distribution is used. 
Setting it to 0 arrives at
\begin{align}
    \hat{\sigma}^{\ast}=\sigma.
\end{align}
\end{proof}

\subsection{The Case of Imperfect Prediction \texorpdfstring{$\hat{\rvp}\neq\bm{\mu}$}{}}
\begin{proof}
TL;DR: when prediction is imperfect, confidence will decrease correspondingly.

It is a more general case and will lead to more misalignment to the ideal score (\Cref{eq:oks_mean}). For instance, the prediction deviation of easy samples is likely to be less than that of hard samples. We derive optimal $\hat{\sigma}$ in this case. It makes sense to some extent for deviation is usually easier to estimate than mean since it only requires to predict a range instead of an exact value. Denote prediction deviation as
\begin{align}
    \hat{\bm{\delta}}=\hat{\rvp}-\bm{\mu},\hat{\Delta}^2=\|\hat{\bm{\delta}}\|^2\neq0;
    \bm{\delta}=\rvp-\bm{\mu}.
\end{align}
For \textbf{Regression}, \Cref{eq:opt_sigma}$=0$ tells
\begin{align}
    {\hat{\sigma}^{\ast}}{}^2&=\frac12{\E_{\rvp}[\|\rvp-\hat{\rvp}\|^2]}=\frac12{\E_{\rvp}[\|\rvp-\bm{\mu}+\bm{\mu}-\hat{\rvp}\|^2]}\\
    &=\frac12\E_{\rvp}[\bm{\delta}^T\bm{\delta}-2\bm{\delta}^T\hat{\bm{\delta}}+\hat{\bm{\delta}}^T\hat{\bm{\delta}}]\\
    &=\frac12(\E_{\rvp}[\|\bm{\delta}\|^2]-2\E_{\rvp}[\bm{\delta}]^T\bm{\hat{\delta}}+\hat{\Delta}^2)\label{eq:delta_const2p}\\
    &=\frac12(2\sigma^2-2\mathbf{0}^T\bm{\hat{\delta}}+\hat{\Delta}^2)=\sigma^2+\frac{\hat{\Delta}^2}2.
\end{align}
\Eqref{eq:delta_const2p} is based on $\hat{\bm{\delta}}$ is constant \wrt $\rvp$. The score \Cref{eq:conf_rle} becomes
\begin{align}
    \hat{\ervs}_{\text{reg}}=1-\hat{\sigma}=1-\sqrt{\sigma^2+\frac{\hat{\Delta}^2}2}<1-\sigma.
\end{align}

For \textbf{Detection}, derivation assumes
\begin{proposition}
Imperfect (but not bad) heatmap follows~\citep{gu2021dive}
\begin{align}
    \hat{\rh}_{\vm}=\hat{\ervo}\exp\left(-\frac{\|\vm-\hat{\rvp}\|^2}{2\hat{\sigma}^2}\right),\label{eq:hm_form}
\end{align}
where $\hat{\ervo}$ is a scaling factor.
\end{proposition}
MSE (\Cref{eq:likelihood}) is derived as
\begin{align}
     \mathcal{L}_{\text{mse}}\overset{c}{=}\E_{\rvp}\left[\sum_{\vm}(\hat{\rh}_{\vm}-\rh_{\vm})^2\right].
\end{align}
For each location $\vm$,
\begin{align}
    \frac{\partial\mathcal{L}}{\partial\hat{\rh}}=\frac{\partial(\hat{\rh}-\rh)^2}{\partial\hat{\rh}}=2(\hat{\rh}-\rh);
\end{align}
\begin{align}
    \frac{\partial\hat{\rh}}{\partial\hat{\sigma}^2}&=\hat{\ervo}\exp\left(-\frac{\|\vm-\hat{\rvp}\|^2}{2\hat{\sigma}^2}\right)\frac{\partial-\frac{\|\vm-\hat{\rvp}\|^2}{2\hat{\sigma}^2}}{\partial\hat{\sigma}^2}
    =\hat{\rh}\frac{\|\vm-\hat{\rvp}\|^2}{2\hat{\sigma}^4},\label{eq:phhat_psigmahat}
\end{align}
\begin{align}
    \frac{\partial\hat{\rh}}{\partial\hat{\ervo}}=\exp\left(-\frac{\|\vm-\hat{\rvp}\|^2}{2\hat{\sigma}^2}\right)=\frac{\hat{\rh}}{\hat{\ervo}}.
\end{align}
Derivation of \Cref{eq:phhat_psigmahat} uses \Cref{eq:pc_psigmahat}.

\paragraph{Detection's Optimal $\hat{\ervo}$} (entangling with $\hat{\sigma}^2$). Further,
\begin{align}
    \frac{\partial\mathcal{L}_{\text{mse}}}{\partial\hat{\ervo}}&=\frac{\partial\E_{\rvp}\left[\sum_{\vm}(\hat{\rh}_{\vm}-\rh_{\vm})^2\right]}{\partial\hat{\ervo}}=\E_{\rvp}\left[\sum_{\vm}\frac{\partial\mathcal{L}_{\vm}}{\partial\hat{\ervo}}\right]\\
    &=\E_{\rvp}\left[\sum_{\vm}\frac{\partial\mathcal{L}_{\vm}}{\partial\hat{\rh}_{\vm}}\frac{\partial\hat{\rh}_{\vm}}{\partial\hat{\ervo}}\right]\\
    &=\E_{\rvp}\left[\sum_{\vm}2(\hat{\rh}_{\vm}-\rh_{\vm})\frac{\hat{\rh}_{\vm}}{\hat{\ervo}}\right]\\
    &=\frac2{\hat{\ervo}}\sum_{\vm}(\hat{\rh}_{\vm}^2-\hat{\rh}_{\vm}\E_{\rvp}[\rh_{\vm}])=0.\label{eq:pmse_pohat}
\end{align}
The last step makes use of that given the image, only $\rh_{\vm}$ depends on $\rvp$.

Denote
\begin{align}
    \hat{\rh}_{\vm}^2&=\hat{\ervo}^2\mathcal{G}_{\vm},\mathcal{G}_{\vm}=\exp\left(-\frac{\|\vm-\hat{\rvp}\|^2}{2\hat{\sigma}^2}\cdot2\right),\\
    \hat{\rh}_{\vm}\E_{\rvp}[\rh_{\vm}]&=\hat{\ervo}\mathcal{H}_{\vm},\\
    \mathcal{H}_{\vm}&=\frac{\tilde{\ervl}^2}{\tilde{\sigma}^2}\exp\left(-\frac{\|\vm-\hat{\rvp}\|^2}{2\hat{\sigma}^2}-\frac{\|\vm-\bm{\mu}\|^2}{2\tilde{\sigma}^2}\right).
\end{align}
$\E_{\rvp}[\rh_{\vm}]$ comes from \Cref{eq:det_score}, and
\begin{align}
    \tilde{\sigma}^2\triangleq\sigma^2+\tilde{\ervl}^2.
\end{align}

Substituting them back to \Cref{eq:pmse_pohat}, we obtain
\begin{align}
    \frac2{\hat{\ervo}}\sum_{\vm}(\hat{\ervo}^2\mathcal{G}_{\vm}-\hat{\ervo}\mathcal{H}_{\vm})&=0\\
    \left(\sum_{\vm}\mathcal{G}_{\vm}\right)\hat{\ervo}-\sum_{\vm}\mathcal{H}_{\vm}&=0\\
    \hat{\ervo}^{\ast}&=\frac{\sum_{\vm}\mathcal{H}_{\vm}}{\sum_{\vm}\mathcal{G}_{\vm}}.\label{eq:ostar}
\end{align}

Consider the limit as $\Delta{\vm}\to\mathbf{0}$ and almost full support of nonnegligible $\mathcal{H}_{\vm},\mathcal{G}_{\vm}$ is within heatmap --
\begin{align}
    \Delta{\vm}\sum_{\vm}\mathcal{G}_{\vm}&\to\int_{\vm}\mathcal{G}_{\vm}\mathbf{d}\vm\label{eq:intro_dm}\\
    &=\int_{\vm}\exp\left(-\frac{\|\vm-\hat{\rvp}\|^2}{2\left(\frac{\hat{\sigma}}{\sqrt{2}}\right)^2}\right)\mathbf{d}\vm=\pi\hat{\sigma}^2,
\end{align}
\begin{align}
    &\Delta{\vm}\sum_{\vm}\mathcal{H}_{\vm}\\
    \to&\int_{\vm}\frac{\tilde{\ervl}^2}{\tilde{\sigma}^2}\exp\left(-\frac{\|\vm-\hat{\rvp}\|^2}{2\hat{\sigma}^2}-\frac{\|\vm-\bm{\mu}\|^2}{2\tilde{\sigma}^2}\right)\mathbf{d}\vm.\label{eq:inth}
\end{align}

Denoting
\begin{align}
    \bar{\sigma}^2\triangleq\tilde{\sigma}^2+\hat{\sigma}^2,
\end{align}
with Lemma~\ref{lem:pphat}, \Cref{eq:inth} is calculated as
\begin{align}
    &\frac{\tilde{\ervl}^2}{\tilde{\sigma}^2}\frac{2\pi\tilde{\sigma}^2\hat{\sigma}^2}{\bar{\sigma}^2}\exp\left(-\frac{\|\hat{\rvp}-\bm{\mu}\|^2}{2\bar{\sigma}^2}\right)\\
    =&\frac{2\pi\tilde{\ervl}^2\hat{\sigma}^2}{\bar{\sigma}^2}\exp\left(-\frac{\|\hat{\rvp}-\bm{\mu}\|^2}{2\bar{\sigma}^2}\right),
\end{align}
\begin{align}
    \hat{\ervo}^{\ast}&\approx\frac{\frac{2\pi\tilde{\ervl}^2\hat{\sigma}^2}{\bar{\sigma}^2}{\exp\left(-\frac{\|\hat{\rvp}-\bm{\mu}\|^2}{2\bar{\sigma}^2}\right)}}{\pi\hat{\sigma}^2}
    =\frac{2\tilde{\ervl}^2}{\bar{\sigma}^2}{\exp\left(-\frac{\hat{\Delta}^2}{2\bar{\sigma}^2}\right)}.\label{eq:opt_ohat}
\end{align}

\paragraph{Remarks of $\hat{\ervo},\hat{\sigma}^2$.} We can further compute
\begin{align}
    \frac{\partial\hat{\ervo}^{\ast}}{\partial\hat{\sigma}^2}=\frac{\partial\hat{\ervo}^{\ast}}{\partial\bar{\sigma}^2}&=-\frac{2\tilde{\ervl}^2}{\bar{\sigma}^4}\exp+\frac{2\tilde{\ervl}^2}{\bar{\sigma}^2}\exp\cdot\frac{\hat{\Delta}^2}{2\bar{\sigma}^4}=0\\
    \text{root }\bar{\sigma}^2&=\frac{\hat{\Delta}^2}2.
\end{align}
Since the derivative is monotonical \wrt $\hat{\sigma}^2$, it is concluded that when $\hat{\sigma}^2>\frac{\hat{\Delta}}2-\tilde{\sigma}^2$, the scale factor $\hat{\ervo}^{\ast}$ decreases with $\hat{\sigma}^2$ (; increases, otherwise).

For \textbf{Detection's Optimal $\hat{\sigma}$},
\begin{align}
    \frac{\partial\mathcal{L}_{\text{mse}}}{\partial\hat{\sigma}^2}&=\frac{\partial\E_{\rvp}\left[\sum_{\vm}(\hat{\rh}_{\vm}-\rh_{\vm})^2\right]}{\partial\hat{\sigma}^2}=\E_{\rvp}\left[\sum_{\vm}\frac{\partial\mathcal{L}_{\vm}}{\partial\hat{\sigma}^2}\right]\\
    &=\E_{\rvp}\left[\sum_{\vm}\frac{\partial\mathcal{L}_{\vm}}{\partial\hat{\rh}_{\vm}}\frac{\partial\hat{\rh}_{\vm}}{\partial\hat{\sigma}^2}\right]\\
    &=\E_{\rvp}\left[\sum_{\vm}2(\hat{\rh}_{\vm}-\rh_{\vm})\frac{\hat{\rh}_{\vm}\|\vm-\hat{\rvp}\|^2}{2\hat{\sigma}^4}\right]\\
    &=\frac1{\hat{\sigma}^4}\sum_{\vm}(\hat{\rh}_{\vm}^2\|\vm-\hat{\rvp}\|^2-\hat{\rh}_{\vm}\E_{\rvp}[\rh_{\vm}]\|\vm-\hat{\rvp}\|^2)\label{eq:pmse_psigmahat}\\
    &=0.
\end{align}
Similarly, we introduce $\Delta{\vm}$ as \Cref{eq:intro_dm}, and the first term becomes
\begin{align}
    \sum_{\vm}\hat{\rh}_{\vm}^2\|\vm-\hat{\rvp}\|^2\Delta{\vm}&\to\hat{\ervo}^2\pi\hat{\sigma}^2\E_{\rvm\sim g}[\|\rvm-\hat{\rvp}\|^2]\\
    &=\hat{\ervo}^2\pi\hat{\sigma}^2\hat{\sigma}^2=\pi\hat{\ervo}^2\hat{\sigma}^4,\label{eq:term1}
\end{align}
as $g(\rvm)=\mathcal{N}(\hat{\rvp},\frac{\hat{\sigma}^2}2\mI)$.

Following Lemma~\ref{lem:pp}, $\mathcal{H}$ can also be expressed as a Normal \wrt $\rvm$ for
\begin{align}
    \exp\left(-\frac{\|\vm-\hat{\rvp}\|^2}{2\hat{\sigma}^2}-\frac{\|\vm-\bm{\mu}\|^2}{2\tilde{\sigma}^2}\right)=\mathcal{K}2\pi\mathcal{J}\mathcal{N}(\overrightarrow{\mathcal{I}},\mathcal{J}\mI),
\end{align}
\begin{align}
    \overrightarrow{\mathcal{I}}=\frac{\tilde{\sigma}^2\hat{\rvp}+\hat{\sigma}^2\bm{\mu}}{\bar{\sigma}^2},
    \mathcal{J}=\frac{\tilde{\sigma}^2\hat{\sigma}^2}{\bar{\sigma}^2},
    \mathcal{K}=\exp\left(-\frac{\hat{\Delta}^2}{2\bar{\sigma}^2}\right).
\end{align}
Thus,
\begin{align}
    &\sum_{\vm}\hat{\rh}_{\vm}\E_{\rvp}[\rh_{\vm}]\|\vm-\hat{\rvp}\|^2\Delta{\vm}\label{eq:raw_term2}\\
    \to&\left(\hat{\ervo}\frac{\tilde{\ervl}^2}{\tilde{\sigma}^2}\right)2\pi\mathcal{JK}\E_{\rvm\sim h}[\|\rvm-\hat{\rvp}\|^2]\\
    \overset{c}{=}&\E[\|\rvm-\overrightarrow{\mathcal{I}}+\overrightarrow{\mathcal{I}}-\hat{\rvp}\|^2]\\
    =&\E[\|\rvm-\overrightarrow{\mathcal{I}}\|^2]+\E[\|\hat{\rvp}-\overrightarrow{\mathcal{I}}\|^2]=2\frac{\tilde{\sigma}^2\hat{\sigma}^2}{\bar{\sigma}^2}+\frac{\hat{\sigma}^4\hat{\Delta}^2}{\bar{\sigma}^4}\\
    \iff&\Cref{eq:raw_term2}=2\pi\hat{\ervo}\tilde{\ervl}^2\hat{\sigma}^4\left(\frac{2\tilde{\sigma}^2}{\bar{\sigma}^4}+\frac{\hat{\sigma}^2\hat{\Delta}^2}{\bar{\sigma}^6}\right)\mathcal{K},\label{eq:term2}
\end{align}
where $h(\rvm)=\mathcal{N}(\overrightarrow{\mathcal{I}},\mathcal{J}\mI)$.

Therefore, substituting \Cref{eq:term1,eq:term2} back into \Cref{eq:pmse_psigmahat} gives
\begin{align}
    \pi\hat{\ervo}^2-2\pi\hat{\ervo}\tilde{\ervl}^2\left(\frac{2\tilde{\sigma}^2}{\bar{\sigma}^4}+\frac{\hat{\sigma}^2\hat{\Delta}^2}{\bar{\sigma}^6}\right)\mathcal{K}&=0
\end{align}
\begin{align}
    \hat{\ervo}&=2\tilde{\ervl}^2\frac{2\tilde{\sigma}^2\bar{\sigma}^2+\hat{\sigma}^2\hat{\Delta}^2}{\bar{\sigma}^6}\exp\left(-\frac{\hat{\Delta}^2}{2\bar{\sigma}^2}\right).\label{eq:opt_sigmahat}
\end{align}

Combining \Cref{eq:opt_ohat,eq:opt_sigmahat} gets
\begin{align}
    \frac1{\bar{\sigma}^2}&=\frac{2\tilde{\sigma}^2\bar{\sigma}^2+\hat{\sigma}^2\hat{\Delta}^2}{\bar{\sigma}^6}
\end{align}
\begin{align}
    (\tilde{\sigma}^2+\hat{\sigma}^2)^2-2\tilde{\sigma}^2(\tilde{\sigma}^2+\hat{\sigma}^2)-\hat{\Delta}^2\hat{\sigma}^2&=0\\
    \hat{\sigma}^4-\hat{\Delta}^2\hat{\sigma}^2-\tilde{\sigma}^4&=0
\end{align}
\begin{align}
    \hat{\sigma}^{\ast}{}^2&=\sqrt{\tilde{\sigma}^4+\frac{\hat{\Delta}^4}4}+\frac{\hat{\Delta}^2}2.
\end{align}

The score is unnormalized density at $\hat{\rvp}$ \Cref{eq:hm_form} as
\begin{align}
    \hat{\ervs}_{\text{det}}=\hat{\ervo}^{\ast}=\frac{2\tilde{\ervl}^2}{\tilde{\sigma}^2+\sqrt{\tilde{\sigma}^4+\frac{\hat{\Delta}^4}4}+\frac{\hat{\Delta}^2}2}<\frac{\tilde{\ervl}^2}{\tilde{\sigma}^2}.
\end{align}
\end{proof}

\subsection{A Misspecification Case of \texorpdfstring{$p(\rvp^{\prime}|\hat{\rvp})\neq\mathcal{N}$}{}}

\begin{proof}
Model misspecification is a common problem in practice. Here, an example of Laplace predictive distribution $\rvp^{\prime}\sim\text{Laplace}$ different from underlying Normal noise distribution is shown. As a result, it still has a perfect prediction $\hat{\rvp}^{\ast}=\bm{\mu}$, but the score deviates from the ideal score (\Cref{eq:oks_mean}).

With Laplace model, \Cref{eq:rle_maximum} is in form of
\begin{align}
    \mathcal{L}_{\text{lap}}&=\E_{\rvp\sim\mathcal{N}(\bm{\mu},\sigma^2)}\left[\log\frac1{4\hat{\ervb}^2}\exp\left(-\frac{\|\rvp-\hat{\rvp}\|_1}{\hat{\ervb}}\right)\right]\\
    &\overset{c}{=}-\E_{\rvp}\left[2\log\hat{\ervb}+\frac{\|\rvp-\hat{\rvp}\|_1}{\hat{\ervb}}\right].
\end{align}
With
\begin{align}
    \frac{\partial2\log\hat{\ervb}}{\partial\hat{\rvp}}=\mathbf{0},\frac{\partial2\log\hat{\ervb}}{\partial\hat{\ervb}}=\frac2{\hat{\ervb}},
\end{align}
\begin{align}
    \frac{\partial\frac{\|\rvp-\hat{\rvp}\|_1}{\hat{\ervb}}}{\partial\hat{\rvp}}=-\frac{\rm{sgn}(\rvp-\hat{\rvp})}{\hat{\ervb}},\frac{\partial\frac{\|\rvp-\hat{\rvp}\|_1}{\hat{\ervb}}}{\partial\hat{\ervb}}=-\frac{\|\rvp-\hat{\rvp}\|_1}{\hat{\ervb}^2},
\end{align}
derivative is computed and set to 0 as
\begin{align}
    \frac{\partial\mathcal{L}_{\text{lap}}}{\partial\hat{\rvp}}&=-\E_{\rvp}\left[\frac{\partial2\log\hat{\ervb}}{\partial\hat{\rvp}}+\frac{\partial\frac{\|\rvp-\hat{\rvp}\|_1}{\hat{\ervb}}}{\partial\hat{\rvp}}\right]\\
    &=-\E_{\rvp}\left[\mathbf{0}-\frac{\rm{sgn}(\rvp-\hat{\rvp})}{\hat{\ervb}}\right]=\E_{\rvp}\left[\frac{\rm{sgn}(\rvp-\hat{\rvp})}{\hat{\ervb}}\right]\\
    &=\frac1{\hat{\ervb}}\E_{\rvp}[\rm{sgn}(\rvp-\hat{\rvp})]=\mathbf{0}.
\end{align}
According to the symmetry of Normal distributions,
\begin{align}
    \hat{\rvp}^{\ast}=\E_{\rvp}[\rvp]=\bm{\mu}.
\end{align}

Besides,
\begin{align}
    \frac{\partial\mathcal{L}_{\text{lap}}}{\partial\hat{\ervb}}&=-\E_{\rvp}\left[\frac2{\hat{\ervb}}-\frac{\|\rvp-\hat{\rvp}\|_1}{\hat{\ervb}^2}\right]=-\frac2{\hat{\ervb}}+\frac{\E_{\rvp}[\|\rvp-\hat{\rvp}\|_1]}{\hat{\ervb}^2}=0
\end{align}
\begin{align}
    \Rightarrow\hat{\ervb}^{\ast}=\frac{\E_{\rvp}[\|\rvp-\bm{\mu}\|_1]}2=\sqrt{\frac2{\pi}}\sigma,
\end{align}
where the expectation is calculated according to $\rvp\sim\mathcal{N}(\bm{\mu},\sigma^2\mI)$.

The score in this case is
\begin{align}
    \hat{\ervs}_{\text{reg}}=1-\hat{\ervb}=1-\sqrt{\frac2{\pi}}\sigma>1-\sigma.
\end{align}
Results may differ from case to case.
\end{proof}

\section{Implementation Details}

\paragraph{Pseudocode} is attached in \Cref{alg:code}, facilitating reproducibility for the community.

\begin{algorithm}[t]
\caption{CCNet Pseudocode, PyTorch-like}
\label{alg:code}
\definecolor{codeblue}{rgb}{0.25,0.5,0.5}
\definecolor{codekw}{rgb}{0.85, 0.18, 0.50}
\lstset{
  backgroundcolor=\color{white},
  basicstyle=\fontsize{7.5pt}{7.5pt}\ttfamily\selectfont,
  columns=fullflexible,
  breaklines=true,
  captionpos=b,
  commentstyle=\fontsize{7.5pt}{7.5pt}\color{codeblue},
  keywordstyle=\fontsize{7.5pt}{7.5pt}\color{codekw},
}
\begin{lstlisting}[language=python]
enc, predhead = freeze(prednet) # the locks in 
 Fig. 2

def forward(x):
    f = enc(x)  # penultimate features
    phat = predhead(f)
    shat, vhat = ccnet(f)
    return phat, [shat, vhat]

def train_step(data, kp_metric=kp_oks, cal_loss=mse, w_vis=2e-2):
    x, p, l, v = data
    phat, [shat, vhat] = forward(x)
    s = kp_metric(phat, p, l)

    loss_kp_conf = cal_loss(shat, s, weight=v)  # calibration loss in Eq. 20
    loss_vis = bce(vhat, v)  # Eq. 21
    loss = loss_kp_oks + w_vis * loss_vis  # Eq. 22
    ...

\end{lstlisting}
\end{algorithm}

\paragraph{Training} cost is minor -- one Fully Connected layer is trained for a few epochs (usually 2 already give good results). Adam~\citep{kingma2014adam} optimizer is used with step Learning Rate decay. Ground truth bounding boxes are provided as input to top-down pose estimation methods.

\paragraph{OKS on MPII~\citep{andriluka20142d}.} Since annotated keypoint sets are different between COCO~\citep{lin2014microsoft} and MPII~\citep{andriluka20142d} but images are similar, per-keypoint falloff coefficients of the neighboring hip, shoulder, and nose are applied to that of the pelvis, thorax, upper neck, and head top, respectively.

\begin{figure}[!t]
    \centering
    \includegraphics[width=0.7\linewidth]{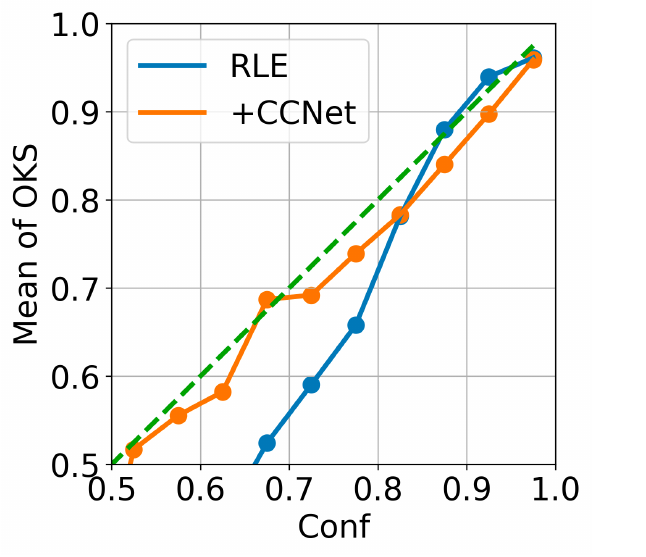}
    \caption{Calibration plot. Estimated confidence well reflects the expected OKS value after pose calibration in our context.}
    \label{fig:ece}
\end{figure}

\paragraph{Pose Calibration}'s variants in \citet{pierzchlewicz2022multi,bramlage2023plausible} are mainly based on keypoint EPE instead of instance OKS for computability and also do not focus on mAP (\Cref{fig:ece}).

\vspace{-1cm}
\vspace{1cm}

\begin{figure*}[!t]
    \centering
    \begin{overpic}[width=0.95\linewidth]{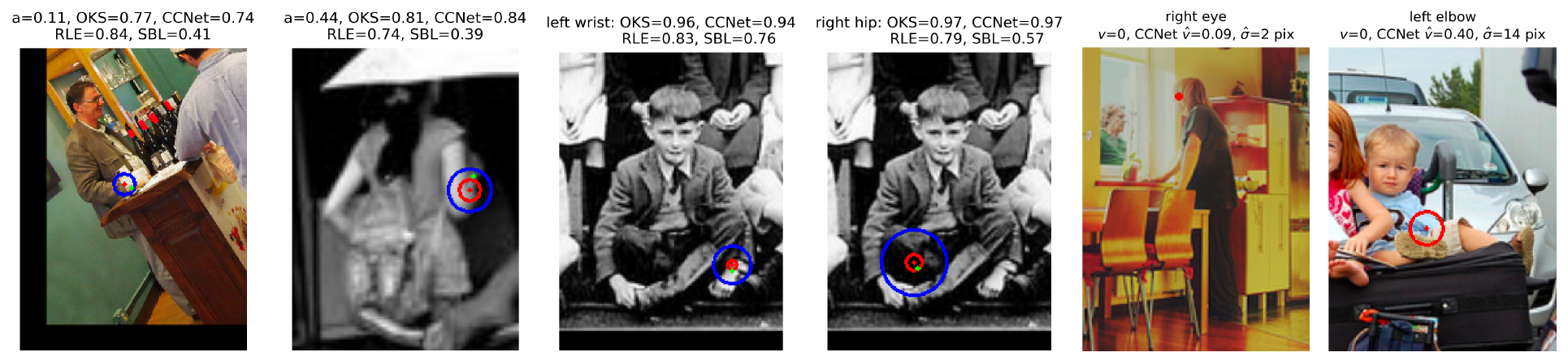}
        \put(16,-2){\small\bf (a)}
        \put(50,-2){\small\bf (b)}
        \put(83,-2){\small\bf (c)}
    \end{overpic}
    \caption{Visualizations of \textbf{(a)} area, \textbf{(b)} per-keypoint falloff (\eg hip's$>$wrist's), and \textbf{(c)} visibility effects to OKS, respectively. {\color{red} Red} dot and circle represent predicted keypoints and sigma confidence; {\color{green} green} dot indicates ground truth keypoint location; {\color{blue} blue} circle depicts OKS $\ervl$ range.}
    \label{fig:viz}
\end{figure*}

\section{More Experimental Results}

\paragraph{Visualizations (\Cref{fig:viz})} show effects of area, per-keypoint falloff, and visibility, respectively. Our CCNet better aligns with OKS. 

\paragraph{Design Choices (Sec.~\ref{sec:design_choices}).} Different losses including Bayesian weight posterior~\citep{lakshminarayanan2017simple} and surrogate optimization~\citep{qi2021stochastic}, perform similarly well (\Cref{tab:losses}).

\begin{table}[t]
    \centering
    \resizebox{\linewidth}{!}{
    \begin{tabular}{lc|cc|c|c}
        \toprule
        & RLE~\citep{li2021human} & Alea~\citep{kendall2017uncertainties} & CE & DeepEns~\citep{lakshminarayanan2017simple} & SOAP~\citep{qi2021stochastic} \\
        \midrule
        {\bf mAP} & 72.2 & 73.4 & 73.5 & 73.5 & 73.4 \\
        \bottomrule
    \end{tabular}}
    \caption{Different loss study other than MSE.}
    \label{tab:losses}
\end{table}

\end{document}